\RequirePackage{rotating} 

\documentclass[manuscript,screen]{acmart}

\AtBeginDocument{%
  \providecommand\BibTeX{{%
    \normalfont B\kern-0.5em{\scshape i\kern-0.25em b}\kern-0.8em\TeX}}}

\setcopyright{acmlicensed}
\copyrightyear{2024}
\acmYear{2024}
\acmDOI{XXXXXXX.XXXXXXX}

\acmJournal{CSUR}
\acmVolume{0}
\acmNumber{0}
\acmArticle{0}
\acmMonth{0}

\usepackage{amsmath}
\usepackage{amsfonts}
\usepackage{graphicx}
\usepackage{wrapfig}
\usepackage[]{natbib}
\usepackage{subfigure}
\usepackage{booktabs}
\usepackage{multirow}
\usepackage{lineno,hyperref}
\usepackage{algorithm}
\usepackage{algpseudocode}
\usepackage{makecell}

\usepackage{float}
\usepackage{verbatim}
\usepackage{pdflscape}
\usepackage{longtable,afterpage}
\usepackage{xcolor}%

\usepackage[small,compact]{titlesec}

\linenumbers

\begin{document}

\title{Humanoid Robots and Humanoid AI: Review, Perspectives and Directions}

\author{Longbing Cao}
\affiliation{%
  \institution{Frontier AI Research Centre, Macquarie University}
  \city{Sydney}
  \state{NSW}
  \country{Australia}
}

\renewcommand{\shortauthors}{Cao}

\begin{abstract}
In the approximately century-long journey of robotics, humanoid robots made their debut around six decades ago. While current humanoids bear human-like appearances, none have embodied true humaneness, remaining distant from achieving human-like to human-level intelligence. The rapid recent advancements in generative AI and (multimodal) large language models have further reignited and escalated interest in humanoids towards real-time, interactive, and multimodal designs and applications, such as fostering humanoid workers, advisers, educators, medical professionals, caregivers, and receptionists. These unveil boundless opportunities of transforming 1) AI robotics into a research era of \textit{humanoid AI}, and 2) AI robots into new-generation \textit{humanoid AI robots} (AI humanoids). Our unique and comprehensive review of about 30 reported humanoids discloses a systematic terminology and a paradigmatic landscape of human-looking to human-like and human-level humanoids. It inspires comprehensive new perspectives and directions of  humanoid AI as an area: transitioning from human-looking to humane humanoids, humanizing humanoids with functional and nonfunctional specifications, and cultivating technical and actionable advances of AI humanoids. Humanoid AI and AI humanoids nurture symbiotic advancements and future opportunities of  synthesizing and transforming humanity modeling and conventional, generative to human-level AI into humanoid robotics.
\end{abstract}




\received{20 February 2007}
\received[revised]{12 March 2009}
\received[accepted]{5 June 2009}

\maketitle

\section{Introduction}
\label{sec:intro}

In the rich history of robot development spanning a century \cite{DarioY17}, \textit{humanoid robots} emerged approximately 60 years ago as a groundbreaking advancement characterizing robots with anthropomorphic forms and appearance \cite{ZlotowskiPYB15}, forming \textit{human-looking humanoids} with some of human structures and appearances. It wasn't until about 30 years ago that humanoid robots began to exhibit notable human-like senses, behaviors, functions, interactions, or reasoning \cite{JAS-Tong23,KajitaHHY14}. This leverages the development of \textit{AI robotics} \cite{murphy2019introduction}, fostering the first generation of AI humanoid robots (or humanoid AI robots, simply \textit{AI humanoids} for short) incorporated with (specific, predefined) capabilities of artificial narrow intelligence (ANI). Such early-stage AI humanoids, like Atlas, Geminoid \cite{Ishiguro2024d}, Pepper and Sophia, were propelled by specific techniques of pattern recognition, computer vision, natural language processing (NLP), signal processing, or shallow machine learning, enabling robot perception, conversation, interaction, collaboration, navigation, or teleoperation. These AI humanoids are typically featured by high-level coarse anthropomorphic structures (e.g. Pepper) with few characterized by static human appearance (like Geminoid and Sophia). They are empowered by on-robot, predefined, unimodal or specific tasks, rules, mechanisms, and functions for given, design-time and pretrained settings and environments. These also triggered a new area \textemdash  \textit{humanoid AI}, marking the dawn of exploratory integration of AI developments into humanoid robotics. 

The recent rapid advancements in deep learning, generative AI (GenAI), artificial general intelligence (AGI), and human-level AI \cite{goertzel2014artificial}, particularly by (multimodal) large language models (LLMs, MLLMs) \cite{MinRSVNSAHR24}, have substantially transformed the area of humanoid AI and the developments of AI humanoids. \textit{Human-like AI humanoids} emerge as AI robots integrated by not only human-looking appearances and structures but also human-like behaviors, facial expressions and even other human qualities, particularly realtime decision-making. AI humanoids, such as the lifelike Ameca \cite{Ameca}, while only a few, have witnessed remarkable progress, showcasing exceptional performance across various aspects and applications \cite{Zeng23}. 

Underpinning AI humanoids is the disciplinary convergence of GenAI, AGI and human-level AI with humanoid robotics, reshaping the area of humanoid AI. \textit{Humanoid AI} migrates into a symbiotic field synthesizing disciplines and areas including robotics, AI, human science, and social science. It transforms AI robots into AI humanoids that are characterized by human-looking and -like structures and functions and are empowered with intelligent capabilities of undertaking both trained and sophisticated untrained tasks at either design or run time and in either static or dynamic settings. Humanoid AI boosts the development of humanoids with multimodal, multilevel, multitask, multiaspect and multipurpose machine intelligence and task capabilities of integrating perception, recognition, conversation, interaction, and operations. Examples of such AI humanoids include Ameca, Phoenix, Optimus, H1, GR-1 and Figure 01 and nonhumanoid robot Google Robot Transformer (RT2). 

Humanoid AI is reshaping the landscape of not only robotics and humanoids but also the intelligent digital economy, societies, and cultures at an unprecedented pace. The rapid developments and evolution of AI humanoids are particularly driven by commercial ambitions, strategies, policies, technology transformation, and investment from government, traditional manufacturing companies, and technology leaders. Projections suggest a substantial growth trajectory of humanoid robots, with the market size of humanoid robots anticipated to reach USD\$38 billion or even USD\$243 billion by 2035, boasting a staggering compound annual growth rate of 13.8-50\% per year according to different market analysts. Meanwhile, GenAI commands an even larger market value, estimated at USD\$2.6-4.4 trillion annually, with a projected 7\% increase in global GDP\footnote{Databricks. The great acceleration: CIO perspectives on generative AI, MIT Technology Review Insights, 2023.}.

Despite these promising projections and fast-paced developments, only a limited number of AI humanoids are empowered by LLMs or GenAI, and none actually hold human-level AI and human intelligence like humanity. This discrepancy is attributed not only to the so-called ``uncanny valley effect'' \cite{MoriMK12} but more fundamentally to the ``humanoid humanity dilemma'' (or more broadly, machine humanity dilemma). \textit{Humanoid humanity dilemma} refers to the plight of empowering human-looking humanoids with realistic to natural humanness. 
Both the uncanny valley effect and the humanoid humanity dilemma highlight significant untapped potential and significant gaps within the realm of humanoid AI, particularly in the foundation and design of realistic humanoids. Accordingly, discussion and speculation about the future trajectory of humanoid AI have to answer a foundational question: \textit{What does the evolution and future hold for AI humanoids}? This has been intensified over the rapid evolution from ANI to AGI or even artificial superintelligence (ASI) \cite{yampolskiy2015artificial} as well as other specialties including metaverse and decentralized AI \cite{CaoDeAI,li2024ugotmeembodiedaffectivehumanrobot}.
\begin{enumerate}
    \item[1)] How will GenAI and AGI propel the evolution of humanoids?
    \item[2)] In what ways will human-level AI redefine the capabilities of  humanoids towards human-level robots with human intelligence?
    \item[3)] How can human cognitive and social features be seamlessly incorporated into physically human-looking robots to enable humane humanoids?
    \item[4)] How will integrating GenAI, AGI, human science and social science into  humanoids set apart humane humanoids from human-looking robots?
    \item[5)] With the advancements in metaverse and decentralized AI, how will humanoid AI integrate computing, learning and services over robot-edge-cloud and metaverse integrative virtual-real-mixed humanoids?
\end{enumerate}

Humanoid AI addressing these issues overturns the current emphasis on constructing humanoid robots with human-like appearances falling short of achieving AGI. To bridge this gap, the integration, simulation, and implementation of human-level AI and human-like intelligence into  humanoids will usher in new eras of human-like and ultimately human-level humanoids. Yet, the development of human-like to human-level humanoids must not only replicate human physical, cognitive, and social attributes but also address ethical, legal, and social concerns and challenges relating to humanity. This suggests \textit{humane humanoids} and humanoid humanity, replicating or simulating human systems, intelligence, quality and humaneness and can accomplish tasks humanly. This requires a revolutionary methodology to drive humanoid AI, such as metasynthetic intelligence and metasynthetic intelligent systems \cite{aikp_Cao15}, and suggests a human-AI-robotics-web-integrative ecosystem of humanoid AI. Going beyond review alone, this motivates our critical and visionary perspectives on 
\begin{enumerate}
    \item[1)] delineating the comprehensive landscape of humanoid robots from those possessing human appearances to early-stage AI humanoids with limited ANI functions and then more realistic AI humanoids driven by GenAI and AGI; 
    \item[2)] articulating a systematic future of humanoid AI and humanoids attaining humane, human-like, or human-level capabilities  embedded with AGI or even ASI. 
    \item[3)] fostering a disciplinary conceptual map and landscape of humanoid AI from functional to nonfunctional specifications and from technical to practical perspectives.
\end{enumerate}    
    
With these objectives and perspectives, our approach focuses not on mechanical, electronic, or biological aspects but rather treats AI humanoids as human-like intelligent systems, with a special emphasis on soft AI-powered humanoids and broadly the area of humanoid AI. We delve into the taxonomy, potential evolution and futures of transitioning physically human-looking humanoids into human-like, and ultimately human-level humane beings, empowered by advancements in ANI, AGI, or ASI for robotics and specifically humanoids. Our reflection of existing humanoids and the big potential nurtures new perspectives and directions of a transformational journey of developing human-like to human-level humanoids. Such advanced humanoids are poised to embody omni-intelligence integrating human-like to -level AI into humanoids and extending the capabilities of humanoids with new paradigms of intelligences, ultimately progressing towards an AGI and ASI-integrated robotic ecosystem. Furthermore, we explore various functional and nonfunctional requirements for human-like to human-level humanoids, as well as techniques and future prospects in modeling deep mind of AI humanoids. We emphasize the importance of enabling and integrating humanity with physical structures of humanoids and cultivating humane humanoids interactive with other objects and humans in humane, dynamic, real-time, and interactive environments. These expect to further advance new-generation humanoids and humanoid AI as a research area.

\section{Paradigms and Categorization of Humanoid AI and AI Humanoids}
\label{sec:humanoidfamily}

Here, we review the paradigm shift of humanoid robots from human-looking humanoids to human-like humanoids. The conceptual ecosystem of humanoid AI is then discussed. Lastly, we present a taxonomy of AI humanoids in terms of both functional and nonfunctional objectives, specifications, tasks and intelligences.

\subsection{Evolution of Humanoid AI and AI Humanoids}
\label{subsec:evolutionstages}

The existing and future humanoid AI research and AI humanoid development can be decomposed and categorized into six aspects and stages with respective focused aims and tasks of 
\begin{enumerate}
    \item[1)] \textit{humanoid structures}: translating human structures into human-looking humanoid structures; 
    \item[2)] \textit{humanoid senses}: translating human senses into human-like humanoid senses; 
    \item[3)] \textit{humanoid behaviors}: translating human behaviors into human-like behaviors; 
    \item[4)] \textit{humanoid functions}: translating human functions into human-like humanoid functions; 
    \item[5)] \textit{humanoid humanity}: translating humanity into human-like humanoid humanity; and 
    \item[6)] \textit{humanoid intelligence}: translating human intelligence into human-level humanoid intelligence, respectively. 
\end{enumerate}

\begin{figure}[t!]
	\centering
	\includegraphics[width=0.8\linewidth]{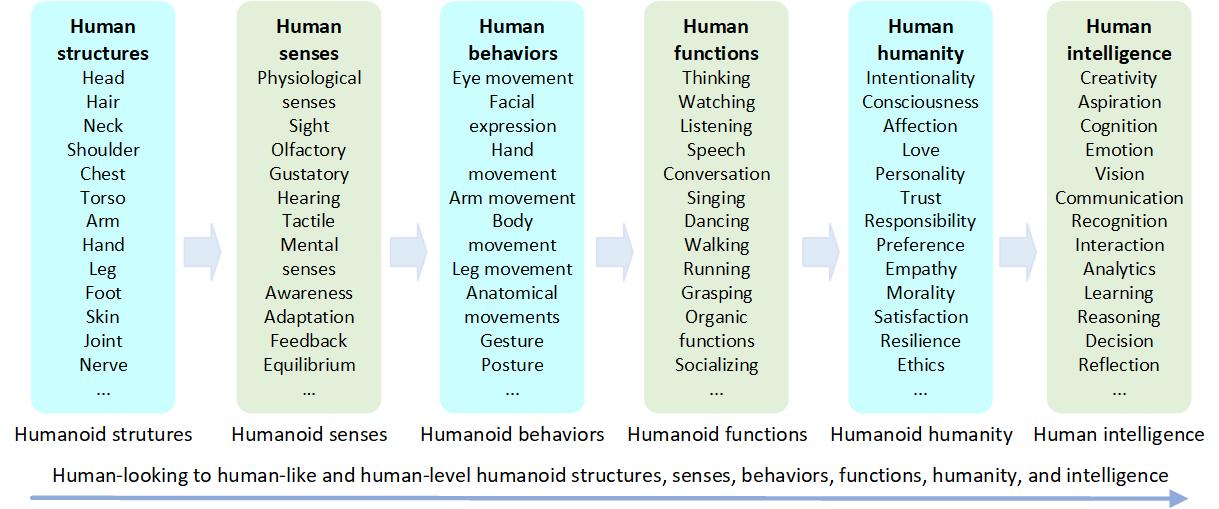}
	\caption{Nonexclusive paradigmatic progression and landscape of human-looking to human-like and human-level humanoid AI research and  AI humanoid developments, resembling human structures, senses, behaviors and functions up to humanity, and intelligence.}
	\label{fig:hrobot-spectrum}
    \vspace{-8pt}
\end{figure}

Fig. \ref{fig:hrobot-spectrum} shows the nonexclusive (with overlap) evolutionary landscape and spectrum of human-looking to human-like and human-level humanoid AI research and AI humanoid developments.
First, the development of \textit{human-looking humanoid structures} aims to replicate human structures and makes robot appearance with \textit{human-looking structures} in part or in full. Typically, a \textit{full-body human-looking humanoid robot} is built with human body structures such as human head, neck, chest, torso, arms, hands, legs and foot. A \textit{half-body humanoid robot} may only have the upper part of human-looking bodies, such as head, neck, shoulder, arm or chest. A humanoid head may be humanly expressive or non-expressive. Most humanoids are incorporated with non-expressive robot heads but with actuated stereo vision, neck or hearing. Few but more modern humanoids are decorated with lifelike expressive android heads, such as with facially expressive traits, artificial skins, neck, or hair.  

Second, the development of \textit{human-like humanoid senses} aims to replicate physiological and mental human senses and develop \textit{human-like senses} in humanoids. Accordingly, such humanoids are built with world perception and sensory modeling. Examples of human-like senses include a visual system for sight, an auditory system for audition, a gustatory system for taste, an olfactory system for smell, an external surface for feel, or cognitive components such as a brain with a vestibular system for balance, orientation, position, depth, and navigation. Most up-to-date humanoids are empowered with vision and audition, none has mental senses implemented. Humanoids like Ameca can express seven basic human feelings and emotions: happiness, sadness, anger, surprise, contempt, fear, and disgust. 

Third, the development of \textit{human-like humanoid behaviors} aims to replicate human behaviors to undertake or generate human-like behaviors or actions in humanoids. Such humanoid behaviors are programmed to humanoid body parts or learned from humans (in person or from video etc) for humanoids to simulate, infer or generate human-like actions. Typical humanoid behaviors include mimicking, inferring or transferring human eye movements, facial expression, and body postures with different hand or leg positions or movements. 

Fourth, the development of \textit{human-like humanoid functions} reproduces human functions of organization, responsiveness and movement in humanoids for them to perform human-like functions individually, in group, or collectively. Typical functions of the human body include speaking, singing, watching, listening, smelling, touching, feeling, grasping, dancing, walking, standing, sitting, and running, etc. These human-like functions are implemented through organizing, responding, moving or developing humanoid structures, senses, and behaviors. Most of humanoids are empowered with part of these functions. While unavailable at the moment, special purpose humanoids may even resemble human functions of metabolism, development, and reproduction.

In addition, the development of \textit{human-level humanoid humanity} aims to make humanoids being humane and having \textit{subjective human features, quality, and states} beyond machines, fostering humanoid humanity. Such \textit{humanoid humanity} may include intention, consciousness, personality, trust, responsibility, preference, empathy, morality, satisfaction, accountability, and ethics. While rarely implemented on existing humanoids, humanoid humanity is essential and critical for humanoids to perform social, cultural, entertaining and industrial activities and interactions in a humane and quality manner.

Lastly, the ultimate goal of humanoid AI and AI humanoids is to develop \textit{human-level humanoid intelligence} by simulating specific to comprehensive, and weak to strong human intelligence into humanoids. Human-level humanoids aim to simulate, infer, reconstruct or generate human-level cognition, vision, emotion and sentiment, communication/conversation, recognition, interaction, collaboration and control, as well as capabilities of analytics, reasoning and inference, learning, decision, and regulation. In general, vision, conversation, recognition, interaction, collaboration, learning and control are not difficult to implement thus widely present in humanoids such as Figure 01 and Optimus. Very few AI humanoids like Ameca can support emotion and sentiment, developing inference and regulation is essential for humane autonomous humanoids. It is an open area to develop humanoid mind though.

\subsection{Humanization Paradigm Shift: Human-looking, Human-like and Human-level Humanoids} 
\label{subsec:paradigm}

Humanoid robots have experienced a fast progression and evolution \footnote{\url{https://en.wikipedia.org/wiki/Humanoid_robot}}.
By reviewing all humanoid robots and their developments over the six categorized  stages of evolution in Fig. \ref{fig:hrobot-spectrum}, humanoid AI studies have experienced a significant paradigm progression over different levels and intensities of humanization. We categorize it into three paradigm stages and aspects:
\begin{enumerate}
    \item[1)] \textit{Human-looking humanoids}, aiming for lookalike humanization with humanoid structures simulating human appearance with human-size and -looking body parts, with or without other human attributes or functions; 
    \item[2)] \textit{Human-like humanoids}, aiming for functional humanization by developing humanoids to replicate human senses, behaviors, and functions; and 
    \item[3)] \textit{Human-level humanoids}, aiming for systematic humanization by nurturing humanoids embodied with substantial and systematic humanity and human intelligence. 
\end{enumerate} 

First, human-looking humanoids are mounted with humanoid structures to make humanoid look like human beings, with some having still or moving attributes. As shown in Table 2 \cite{HumanoidAI25}, most of humanoids are equipped with simple and abstract human-looking parts, while robots like Geminoid and Junko Chihira are with more nuanced facial and head features. Humanoids may be driven by wheel and bipedal facilities. Then, human-like humanoids further humanize robots with representative human traits and functions. Humanoid senses, behaviors and functions make human-like humanoids not only look humanly but also behave and function humanly. Such humanoids tend to support human-friendly immersive social interactions, such as by nuanced eye contact, and expressive facial and and postural response. Lastly, human-level humanoids narrow the gaps between humanoids being machine with machine intelligence and the clone (`doppelganger') of humans with some of simulated human uniqueness and identity of humanity and intelligence.

A humanoid may blend humanoid structures with some of humanoid senses, behaviors and functions and specific humanoid intelligence suiting the humanoid objectives, tasks, and settings. More advanced human-looking robots tend to also embrace more expressive humanoid behaviors, such as body posture and actions, and even facial expressions. Most advanced AI humanoids may have electronic and AI-empowered brain, hands and legs and pursue humanoid functions such as conversation, speech, singing, dancing, walking, running, jumping, or grasping. 

There are about 30 types of humanoid robots available in the market and literature, while this number is rapidly updated. Table 2 in \cite{HumanoidAI25} provides an overview and comparison of 30 humanoid robots identifiable in public resources, which are endowed with certain human-like features and technical and AI-enabled functions. From the AI capacity perspective, we categorize existing humanoids into 1) \textit{naive human-looking humanoids} with some human appearance but limited to none AI functions, 2) \textit{ANI-driven standalone humanoids} with limited ANI functions on board only, and 3) \textit{GenAI-enabled networked humanoids} with on-humanoid and cloud-based operations and more capable AI functions in particular GenAI capabilities like Ameca with real-time connection to ChatGPT. Humanoids at these three capacity phases do not reach AGI capabilities on humanoid or on cloud.

\subsection{Ecosystem and Taxonomy of Humanoid AI and AI Humanoids} 
\label{subsec:ecosystem}

The six evolutionary stages of AI humanoid development and the three paradigms of humanoid AI research 
lay the foundation of the ecosystem, taxonomy and applications of humanoid AI and AI humanoids. Below presents a unique, multi-dimensional, multi-view and evolutionary taxonomy of humanoid robotics, humanoid AI, and AI humanoids.

\subsubsection{Humanoid AI As A Field with Human-AI-Robotics-Web-Integrative Ecosystems}
As a new research area, humanoid AI surpasses AI and robotics and transcends specific progresses such as GenAI, LLMs, and LMMs which are intensively applied in AI humanoids. Humanoid AI has ushered in a new era with revolutionary advancements and possibilities: 
1) Transitioning from traditional task-specific hand engineering and programming methodologies to semi-task-specific, task-agnostic, or even open-task design, development, evaluation and deployment; fostering greater functional and task versatility and flexibility of humanoids.
2) Evolving from predefined, design-based and rule-driven robotic functioning, behaving and servicing to online, real-time, and active learning-driven tasking,  execution and adaption; continuously improving humanoid performance, adaptability, real-world feedback, and experience.
3) Upgrading from individual robot-centric approaches to multi-robot, multi-task, multi-party, and entire process-oriented operations, control, planning, and task execution; enhancing humanoid efficiency and scalability in complex tasks, settings and environments.
4) Progressing design and development objectives from functional to nonfunctional and combining humanization in humanoid studies and system design; pursuing assurance of humanoid performance, quality and services in humane and natural settings and environments.

The six evolutionary stages in Section \ref{subsec:evolutionstages} and the three paradigms in Section \ref{subsec:paradigm} address the above transitions and progressions of humanizing humanoids and developing humanoid AI over phase. They lay the foundation and sketch the scope of the area of humanoid AI. Building on and beyond individual areas and AI robotics \cite{murphy2019introduction}, \textit{humanoid AI} emerges as the disciplinary synergy and metasynthesis \cite{aikp_Cao15} of robotics, AI, human science, and social science, specifically human systems, and intelligent systems, presenting as a human-AI-robotics-web-integrative ecosystem.

\subsubsection{Taxonomy of AI Humanoids}
\begin{figure}[t!]
	\centering
	\includegraphics[width=0.9\linewidth]{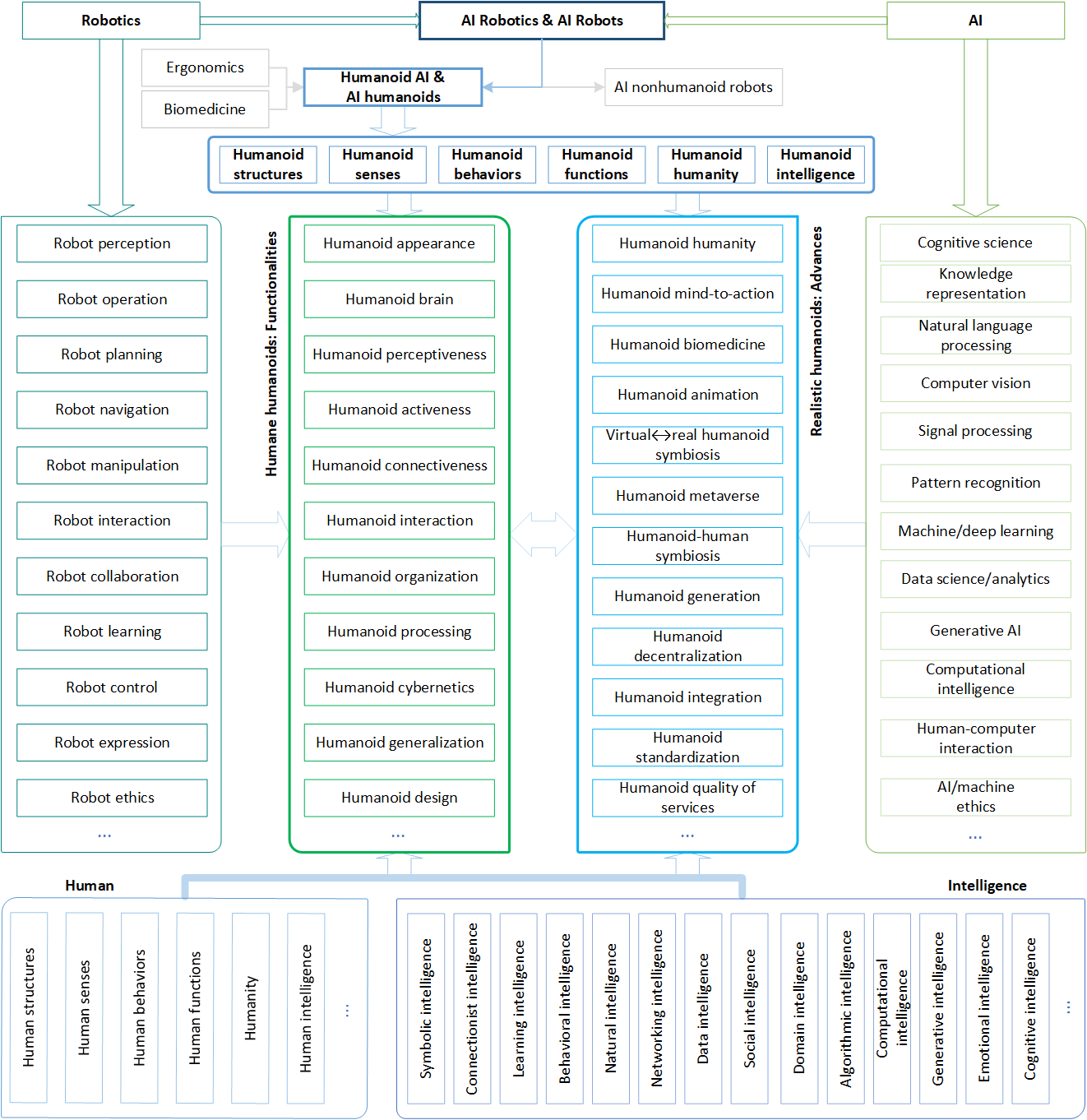}
	\caption{The taxonomy of humanoid AI and AI humanoids: Symbiosis of AI, robotics, human science, and intelligence science.}
	\label{fig:hrobot-family}
    \vspace{-12pt}
\end{figure}

With the ecosystem of humanoid AI, we further explore the interactions and synergy between AI and robotics and between human science and intelligent science. They result in the taxonomy of humanoid AI and AI humanoids, as shown in Fig. \ref{fig:hrobot-family}. 

First, on the high and technical level, \textit{AI robotics} \cite{murphy2019introduction} is the synergy of robotics and AI. In AI robotics, AI functions, tasks and techniques make robots more intelligent, enabled by cognitive science, computer vision, NLP, speech recognition, signal recognition, pattern recognition, machine and deep learning, data analytics, computational intelligence, knowledge representation, human-computer interaction, and machine ethics. AI robotics expand robotic functions, tasks, and techniques from replication to learning and generation.   

Second, \textit{AI humanoid robotics}  incorporates humanoid features into AI robotics, producing AI humanoids and the field of humanoid AI. The foundations and enabling technologies of humanoid AI include conventional, general, generative, decentralized and emotional AI, as well as safe, ethical and responsible AI. Specialties include broad AI techniques, such as robotics, computer vision and pattern recognition, NLP/LLM, reinforcement learning, machine learning, deep learning, data science, and business informatics, human-machine interaction, and human and social science.  

Then, on the methodological level, AI humanoids incorporate elements and systems in human science and intelligent science respectively and integrate them into cross-disciplinary unification and transformation towards AI humanoid robotics. \textit{Human science} studies important traits and merits, such as human structures, senses, behaviors, functions, humanity, and intelligence. Incorporating such human systems into AI humanoids makes humanoids human alike and humanized. Further, \textit{intelligent science} have or implement various paradigms of intelligences. Such X-intelligences \cite{Cao22b} may traverse traditional intelligence paradigms such as symbolic, connectionist, natural, social, domain, computational and networking intelligence to emerging paradigms including data, algorithmic, learning, generative, behavioral, emotion and cognitive intelligence.

Further, \textit{AI humanoid capabilities} comprise both functional and nonfunctional aspects to implement humanoid senses, behaviors, functions, and nonfunctionalities. Functionally, AI humanoids interact, operate, communicate, coordinate, manipulate, navigate, automate, generate, simulate, replicate, generate, synthesize or meta-synthesize \cite{aikp_Cao15} traits and merits of human systems and intelligent systems. Nonfunctionally, AI humanoids may learn to empathize, interpret, secure, trust, derisk and comply. These functional and nonfunctional aspects expand, upgrade and transform robots toward a spectrum of humanoid/humanized functions. 

Accordingly, a family of \textit{research areas} of humanoid AI emerge, such as humanoid perception, planning, cognition, analytics, learning, reasoning, inference, reinforcement, reconstruction, generation, navigation,  manipulation, control, regulation, compliance. Multi-humanoids may develop capabilities for communication, interaction, collaboration, and  transfer. Humanity may be built into humanoids, such as emotion, intention and empathy. These humanoid AI research may apply to individual or group humanoids and for developing human-looking, -like and -level humanoids. 

\subsubsection{Applications of Humanoid AI and AI Humanoids}
The applications of humanoid AI address various purposes  and generate diversified types of AI humanoids. Functional humanoids can be categorized into social humanoids, industrial humanoids, and research humanoids. Their functional examples include walking, conversational, interactive, expressive, generative, manipulative, teleoperated, cognitive, and imitative humanoids. With regard to domains, AI humanoids are increasingly applied to widespread and specified situations and environments  \cite{KajitaHHY14},  professionalized into service, entertaining, convention, caregiving, assistant, advising, home, factory, manufacturing, logistic, workflow, online, arts, military, search, exploratory, security, compliance and social humanoids, and other task-specific humanoids. 

Examples of specific humanoid applications include \textit{home, residential, personal and elderly assistance and caregiving humanoids} for healthcare, home care, and aged care; \textit{safety humanoids} for work, road, transport and traffic safety detection, monitoring and assurance; \textit{social humanoids} for social services, activities and engagement; \textit{education humanoids} such as for primary education, inquiries, question/answering, and online learning; \textit{marketing and customer services humanoids}, such as for general inquiries, reception, chatbot, campaign, and activity organization and management; \textit{hospitality humanoids} such as for kitchen, food and catering services; \textit{digital art humanoids} such as for creating digital songs, dances and artworks; \textit{health and medical research and service humanoids} such as for prosthesis, orthosis, personalized healthcare aids, eldercare, companionship and nursing support, and supporting people with disabilities; \textit{manufacturing and industrial humanoids} such as for mining, assembly, sorting, painting, material handling, pick-and-place, and delivery, and operating specially designed equipment and vehicles, such as stuntronics and animatronic devices; \textit{cognition and neuroscience research humanoids} such as for human brain modeling and cognitive functions on neurorobots; \textit{military humanoids} for unmanned military agents, operations, and activities; \textit{exploration humanoids} such as for astronomy and space exploration; \textit{emergency humanoids} such as for disaster management and performing hazardous jobs such as nuclear and earthquake rescue; \textit{event, demonstration and exhibition humanoids} for their instruction, planning, coordination, monitoring, demonstration, and large-scale social activities such as for theme parks and exhibitions; and \textit{entertainment, game and competition humanoids} such as humanoid league for RoboCup
and World Humanoid Robot Sports Games for half-marathon competition.

\section{Humane Humanoids: Functional and Nonfunctional Specifications}
\label{sec:requirements}

\begin{sidewaysfigure}
    \centering
    \includegraphics[width=1.0\linewidth]{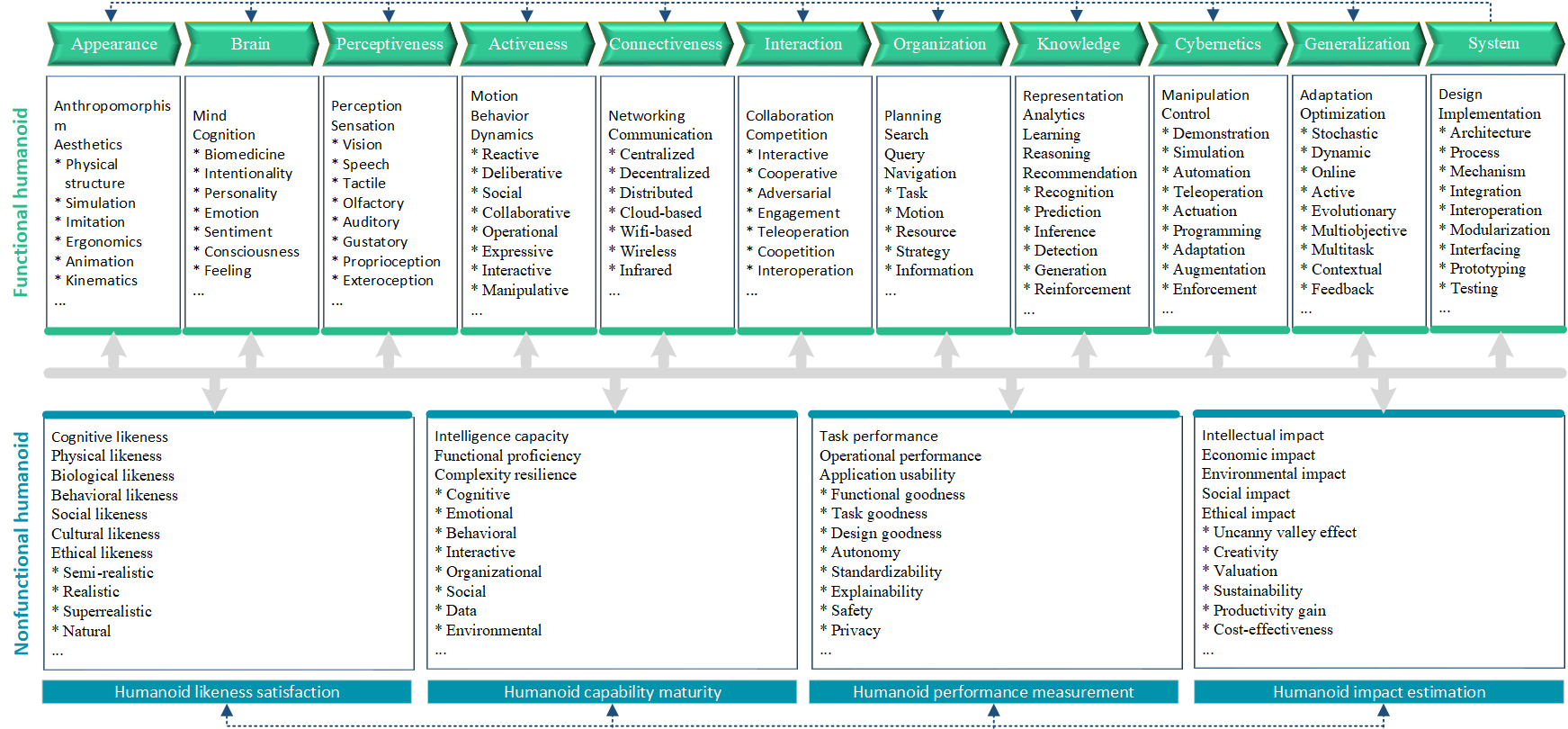}
    \caption{Humane humanoids with functional and nonfunctional specifications of humanizing human-looking to human-like and human-level humanoids.}
	\label{fig:Humanoid_robot_intelligence}
\end{sidewaysfigure}

Humanoid AI develops AI-driven humane humanoids sufficing functional and nonfunctional specifications with corresponding criteria, constraints, and quality measures in performing their objectives, designs, tasks, and properties. This is applicable to human-looking to human-like and human-level paradigms and generations of humanoid AI and AI humanoids. Fig. \ref{fig:Humanoid_robot_intelligence} summarizes typical functional and nonfunctional specifications of AI humanoids.

\subsection{Functional Specifications of Humanoids}
\label{subsec:functional}

Functional specifications of human-like and lifelike humanoids define and enable how an AI humanoid look like, what it does, and how it implements its purposes, tasks, and operations, etc. They determine what humanoid structures, senses, behaviors and functions to implement and operate, as summarized in Table 3 \cite{HumanoidAI25}. While humanoids differ in their functions and none holds all functions, humanoids are expected to satisfy functional specifications defining  mechanical, electrical, biological and social designs with anthropomorphic and mimetic features. Such humanoids imitate human intelligence by their enabling appearance, brain, perceptiveness, activeness, connectiveness, interaction, organization, knowledge, cybernetics, generalization, and system.

\subsubsection{Humanoid Appearance: Anthropomorphism and Aesthetics}
Anthropomorphism and aesthetics customize humanoid appearance. Humanoids are partially to fully embodied with anthropomorphic, humanly mimetic, morphological, animated or caricatured features. This humanoid anthropomorphism  \cite{ZlotowskiPYB15} and aesthetics  include instantiations into: 1) \textit{human-like or lifelike organs}, such as electronic skin (e-skin), bionic and sensory tactile skin, prosthesis, wearables, and 3D surfaces; and 2) \textit{human-like or lifelike functions, senses and traits} embedded into anthropomorphic body parts, such as touch, pressure, force, contact, hardness and texture, or even humanly physiological features such as temperature, pain, pulse rate, blood pressure, and bodily fluids such as sweat and tears. Such \textit{human mimetic mechanisms} \cite{Asano17} and anthropomorphism further evolve into \textit{human imitation}, embedding humanoids with human-like and lifelike body structures and functions, such as skeletal structures and functions, human body proportions, link length, mass balance, muscle arrangement, insertion points, and joint performance with joint range and output power; \textit{anthropomorphic behaviors and activities} such as eye movement, head movement, skin sense, tactile sense, and hand gesture; and \textit{human-like and lifelike body quality}, such as degrees of freedom, expression intensity, and sense sensitivity.  

Transferring human-like and lifelike features into humanoids fosters humanoid aesthetics. \textit{Humanoid aesthetics} synthesize human beauty, creation art, fashion, and visualization into humanoid beauty, typically embodied through avatars. For example, cybernetic avatars \cite{Ishiguro2024d} may have very nuanced positions, shapes, colors and degrees of freedom of body parts, such as realistic facial skins and contours, and eyes with fine-designed eyelids, eyeballs, and eyebrows. Technological aesthetics may produce realistic to hyper-realistic humanoids, the alter ego of humans. Such near-perfect humanoids and avatars may trigger an experience of the uncanny to some people \cite{MoriMK12}, also because of their lack of the nature of human mind, personhood, and humanity. Further developing digital mind and humanity, such as mental states, emotions, intention and ethics, into humanoids may leverage humanoid anthropomorphism and aesthetics with more humane characters, as further discussed in Sections \ref{subsec:nonfunctional} and \ref{subsec:humanity-subjectivity}.

\subsubsection{Humanoid Brain: Mind and Cognition}
\label{subsubsec:brain}
Humanoid mind and cognition imitate human brain mechanisms to achieve human-like mental and cognitive thinking, consciousness, traits, and capabilities, etc., although rarely explored. Accordingly, developing humanoid mind and cognition requires to understand how human brain functions and to develop humanoid mental models simulating human functions and capabilities including reasoning, inference and decision-making mechanisms, as well as mentality like goal, intention, and emotion. \textit{Humanoid mental modeling} attributes the theory of mind with thoughts, desires and intentions etc. \cite{Tabrez2020ASO}. It formalizes mental representations shareable between humanoids and between humans and humanoids by characterizing and simulating the awareness, roles, responsibilities, knowledge structures, mechanisms and processes of human cognition, reasoning, teaming, interaction, communication and collaboration for achieving human-like problem-solving and decision-making. 

Specifically, \textit{humanoid intentionality} captures how humans make aims, objectives, and plans and develops robotic intention models. \textit{Robotic emotions} are implemented by hand-coded, automated, or learning-based designs and methods \cite{RawalS22}. \textit{Robotic affection} further combines personality trait, attitude, emotion, mood and interpersonal stance \cite{McCollHHNB16}. Modeling humanoid intention, emotion and affection represents a new stage of developing humane humanoids by incorporating humanity and mentality into human-like, human-level machines. Humanity and mentality may be learned from human activities, scenes, behaviors, facial expressions, spoken emotions, postures, textual sentiments, and interaction attributes, etc. The generation, recognition, learning, transfer, alignment and adjustment of humanoid emotion, affection and intention may be guided by goals and tasks, in the wild, during performing activities, from multiple modalities, or during interaction and collaboration with humans.

\subsubsection{Humanoid Perceptiveness: Vision, Speech, Hearing and Senses}
\textit{Humanoid perception} simulates human sensing mechanisms of vision, speech, hearing, touch, taste, smell, and other sensations such as feeling. Humanoids perceive, represent, understand, recognize, fuse, adapt and respond to internal and contextual information, activities, and situations through sight, oral, auditory, tactile, olfactory, gustatory, and other sensory channels. Humanoids are equipped with proprioceptive and exteroceptive sensors to perceive and model internal states and behaviors, target objects and their states and behaviors, and interactions with objects w/wo context or environment. Internal states of humanoids may include information about their locomotion, position, velocity, depth, orientation and behaviors \cite{Roychoudhury23}. Humane humanoids may also perceive cognitive features such as attention, gaze movement, and facial expression or even emotion of a human. They also identify and recognize  target objects and their layouts, states, behaviors, or activities, e.g., human pose and facial features. During humanoid-object interactions and human-humanoid interactions, humanoid further detects, recognizes and reacts to interaction modes, activities, types, modalities, and channels. Contextual perception captures the context, surroundings or environment of a humanoid and its objects. Humanoid perceptron identifies contextual materials and textiles, environmental states, conditions (e.g., lightening conditions), or layouts. With embedded cameras to enable robot vision \cite{Robinson23}, \textit{humanoid vision} can capture and enable humanoid gaze, gesture, pose, action, motion, expression, movement, collaboration, learning, and communication. 

Humanoid perception \cite{Roychoudhury23}  is enabled by either \textit{intrusive sensors} such as actuated cameras, force, microphone, inertial measurement units, encoders, sensitive resistors, sonar, rotary, tactile, torque, laser scanner, joint and inclinometer, which may be placed in a humanoid's head, neck, chest, torso, or hand etc body parts, or \textit{non-intrusive sensors} or facilities such as remote wireless cameras, radar, LiDAR, or Wifi. Sensors or multi-sensor fusion may capture internal and external states, signals or videos relating to vision, audio, olfactory, tactile, shape, curve, and posture of human body parts; human features such as emotion and facial expression; physiological signals of human body, such as heart rate, blood pressure, body temperature, brain activity, and muscle activation; and environmental range and layout, or contextual representation and memory, etc.  In performing perception, humanoids recognize internal states such as stability, dynamics, safety, efficiency, and experience; and external aspects such as identifying, recognizing and locating external stimuli, layout and relationships with objects; and interactive states such as robustness and adaptability to dynamic and unforeseen environmental changes.
Humanoid perception enables diverse humanoid tasking and functions like tracking, detection, recognition, identification, classification, control, localization, navigation, planning, mapping, manipulation, and grasping. Humanoid perception can also support specific tasks relating to human-humanoid interactions, object manipulation, gait planning, proprioceptive state estimation, action prediction, and emotion recognition.

\subsubsection{Humanoid Activeness: Behaviors, Motions and Dynamics}
\label{subsubsec:activeness}
Humanoid activeness are embodied through their behaviors, kinematics, and dynamics. The dynamics of human systems, behaviors, and intelligence are fundamental for human functions, operations, and evolution. Understanding human behaviors and dynamics can inform and transform humanoid behaviors and dynamics. 
\textit{Humanoid behaviors} capture how a humanoid behaves, acts, and reacts, etc., supporting various forms of operations, such as gesture, movement, and interactions with other robots or humans \cite{Barmann23}. Behaviors emerge over the lifecycle of humanoid tasking, generating sequential and full-cycle state spaces with behavior attributes, processes, consequences, and impacts \cite{Cao10}. For example, \textit{humanoid gestures} \cite{CarfiM23} comprise human gesture-simulated appearances, behaviors, and states, enabling humanoid's facial expression, eye gaze, communicative pointing and signs, and manipulative motions, etc. Humanoid movements embody actions such as eye, hand, and leg movement. These low-level behaviors further enable high-level \textit{humanoid motions}, implementing general behavioral functions and purposes such as communication, engagement, grasping, navigation, and teleoperation. Under specific tasks and scenarios, humanoids may take imitative, tactile, social, walking, grasping,  acrobatic, or amphibious behaviors, which may be stylistic, personalized, safe or hazardous.

\textit{Humanoid navigation} goes beyond robot navigation \cite{MollerFBHF21} of taking actions and following action trajectories to achieve goals and intention. In general, humanoids navigate passively, guided by visual, vocal or other types of perception, a map, demonstrations, simulation, or pretraining. 
Humanoids manipulate their gestures, movements, and actions to change and generate new dynamics \cite{SugiharaM20}, further driving mobility and dynamics of their kinematics, contact mechanics, and centroidal dynamics with bipedal or wheeled devices. Increasingly, humanoid \textit{learning to navigate} or \textit{navigation by learning} drives active, online, mapless or open-world exploration, learning, and reinforcement \cite{WangZ24}, enabled by unsupervised, active, online, adaptive, reinforced, and open-world learning. Humanoid learning-to-navigate capabilities handle not only general robotic but humanoid-specific tasks and settings, such as navigating while assisting elderly people for humanoid health care, which may fuse special navigation specifications (e.g., safety and trust check). 

\textit{Humanoid teleoperation} traverses robot teleoperations \cite{DarvishPRCPYIP23} of undertaking local behaviors and telebehaviors, such as telelocalization, teleperception, retargeting, mapping, planning, and control. The human-like nature makes humanoid more suitable for teletasking, telepresence, and teleexistence in remote environments, such as for telehealth and telenursing. Humanoids may be teleoperated in a whole-body, partial-body, bilateral, leader-follower, stereoscopic or even immersive manner, where humanoids may be manipulated by speech, position force control, eye gaze, bimmanual control, telelocomotion, imitation, demonstration, learning, or even virtual/mixed reality \cite{PencoMMAKCG24}. As discussed in Section \ref{subsec:metaverse-humanoid}, humanoid teleoperation can further integrate animation, virtual reality, digital twin, and metaverse technology for more immersive and realistic operations. In this regard, a typical example is the \textit{cybernetic avatar} \cite{Ishiguro2024d}, teleoperated by an operator, where the avatar and the operator coexist as a symbiosis. 

In addition, humanoid behavior cloning, imitation, animation, transformation, adaptation, and transfer can mimic, share, transform, generate or respond with new behaviors, fostering new humanoid functions and operations. These would further involve humanoid behavior analytics and computing \cite{Cao10} to understand, monitor, predict, and manage behaviors and their effects and improve humanoid goals, tasks, and performance.

\subsubsection{Humanoid Connectiveness: Networking and Communication}
\label{subsubsec:connectiveness}
Humanoid connectiveness builds on humanoid networking and communication \cite{NortonACFGGSSSS22,Juang24}. Humanoid networking builds connections within or between humanoids and between humanoids and external objects, humans and environments. Humanoid communication further imparts, transports, or exchanges information, materials, or energy within a humanoid, between humanoids, between a humanoid and its environment, or between humanoids and other objects such as humans. \textit{Learning to communicate} (LoC) rapidly overtakes roles of traditional communication methods, physical objects, and signals. LoC increasingly involves demonstrations, imitations, representations, or model parameters of physical or sensory materials, behaviors, or activities. Besides \textit{communication participants} like other humanoids, humans, or objects, LoC can involve \textit{media} such as in imagery, video-based, acoustic, textual, emotional, or behavioral form; \textit{content} such as speech, navigation, and actions; \textit{indirect participants} such as humans or objects for demonstrations or imitations; and \textit{context} such as topic areas and background. While biparty communications with another party like a human or another robot dominate existing studies, peer-to-multiparty interactions represent frontiers by enabling multiple parties, or a team with multiple humanoids, or humans and humanoids. Such human-to-humanoid communications can be unidirectional, directional, or multidirectional.
Humanoid communications happen 1) within humanoids in various ways, e.g., for body part coordination between head and hands or between eye and hands through transiting and processing sensory information, and state or process transition or transformation such as from perception representation to conversation; 2) between humanoids or between humans and robots, e.g., between humanoids for collective tasks through goal sharing, task cooperation, or load negotiation; and 3) between humanoids and their static or dynamic contexts/environments, e.g., through tracking, representing, and transiting contextual information for navigation and operations; etc. 

LoC-driven humanoid communications expand \textit{linguistic} communications from speech, message-passing, and dialogue to query-answering, retrieval, and recommendations; and \textit{non-linguistic} communications from vision \cite{Juang24} and camera/eye contact to responsive emotions, and postural, gestural, behavioral, or graphical interactions or movements. These transform unimodal to multimodal LoC with language-based to multimodal communications. \textit{Non-expressive} communications such as through sensory information consumption and sharing are complemented by \textit{expressive} communications such as emotional, sentimental, affective, or behavioral expressions and exchange of facial expressions, spoken emotions, textual sentiments, or behaviors such as gestures, postures, body language, and actions. Besides multimodalities, \textit{multimodal LoC} also incorporates \textit{multimodal transformation and translation} in \textit{bilateral} manners, such as via text-to-speech translation, speech-to-text translation, image-to-text transformation, text-to-image transformation, vision-to-language transformation, and language-to-vision transformation;  \textit{multilateral}, such as vision-language-to-action (VLA) transformation, or vision-language-to-emotion (VLE) transformation \cite{li2024ugotmeembodiedaffectivehumanrobot}; or \textit{mechanism-based}, e.g., demonstration-to-action, or imitation-to-action. These translation and transformation tasks undertake cross-modality or cross-mechanism information transport, which may take place in an \textit{invasive}, \textit{non-invasive}, or hybrid mode.

\subsubsection{Humanoid Interaction: Collaboration and Competition}
\label{subsubsec:interaction}
Humanoids connect, associate or relate one to another or to other objects or humans, forming humanoid interactions for collaboration or competition. Such interactions take place between humanoids, humans, or environments, and within humanoid communities, transforming human-robot interactions (HRI) \cite{GoodrichS07} to specific humanoid-humanoid, human-humanoid, or human-humanoid-environment interactions for humanoids. \textit{Humanoid interactions} may be directional, physical, cognitive, perceptual, behavioral, responsive, social, multi-party-based, contextual, or environmental, depending on their interaction settings, environments, and tasks. Specifically, humanoid interactions transcend general forms of robot and human-robot association, connection, recognition, identification, detection, or tracking to specific forms of humanoid or human-humanoid coordination, cooperation, negotiation, collaboration, or even conflicting, or competition. Humanoid interactions are associated with, during or for implementing tasks, such as grasping, assembly, communications, collaboration, or teleoperation. For example, \textit{humanoid-human collaboration} builds on robot collaboration or human-robot collaboration \cite{BauerWB08,SemeraroGC23} to fulfill tasks including forming teams, defining goal, plan, actions, strategies, and constraints, resolving conflicts, or enabling humanoid communications, analytics, and learning. With humanoid learning, \textit{learning to collaborate} grows its value in humanoid collaboration, empowering humanoids to \textit{learn from data} such as hand-crafted knowledge base, labeled samples, demonstrations, imitations, machine-generated samples, or protocols; and to learn by methods like pretraining, retraining, finetuning, or adaptive or continual learning. Learning from mixed reality and metaverse could also transform humanoid interactions in physical worlds. In contrast, \textit{humanoid competition} may be driven by goals, tasks, skills, resources, roles or scenarios. Adversarial competition may involve humanoid attack and defense. 

\textit{Social humanoids} engage \textit{social interactions} \cite{Korn2019} in a humanoid society, agency, or human-humanoid coexisting community. Social humanoids follow social and cultural norms and express emotions and sentiments during social engagement and in fulfilling social skills and solving social problems. Social interactions incorporate \textit{humanoid humanity}, such as intention, emotion, affection, and empathy. For example, affective robot interactions \cite{OlugbadeHMHB23} may enable better collaboration, assistance, imitation, coordination, cooperation, and general or multi-purpose communications.

\subsubsection{Humanoid Organization: Planning and Search}
Humanoid organization decides what, how, when, or where a humanoid behaves, undertaking robot planning and search \cite{VikasP23} to navigate, move body parts, search information, take actions, or perform tasks. These enable path, motion \cite{TazakiM20}, and task planning, or incorporate robot search, such as heuristic search, and stochastic search, into retrieving interested information or target, in known to unknown clutter environments with known to unknown obstacles. In doing these, humanoids optimize their target motions, movements, paths, or tasks and minimize concerning issues such as collisions, and time or energy consumption.

With humanoids enabled by learning advances including LLMs and LMMs, \textit{learn-to-plan}, or \textit{planning by learning} \cite{GanesanKCM24}, becomes a new fashion of humanoid planning in complex contexts. Humanoids navigate, search, respond and act per justifications, predictions, recommendations, or retrieved findings by techniques including probabilistic roadmap, evolutionary computing, geometric modeling, reinforcement learning, and deep learning. Learn-to-plan is particularly important for interacting with, manipulating, or responding to unseen objects, scenarios, tasks in semi-open to open worlds. For example, a robot enabled by a pretrained VLM can extract object category-specific features from images and texts \cite{Stone-23}. Further, a robot can undertake actions or complete instructions with policies informed or constrained by categorized objects with imagery and textual instructions, such as SayCan \cite{IchterBCFHHHIIJ22} which \textit{plans-to-act} based on input language. Google RT-2 \cite{GoogleRT-2} is another VLM-based learn-to-plan advance, which builds on a VLM backbone, plans from both image and text commands, and conducts visually grounded planning. Its \textit{chain-of-thought} reasoning can enable learning long-horizon planning and low-level skills to achieve highly-improved robotic policies by instantiating VLAs based on PaLM-E \cite{DriessXSLCIWTVY23} and PaLI-X \cite{Chen0CPPSGGMB0P23}. RT-2 can significantly improve generalization performance and emergent capabilities using the power of web-scale \textit{vision-language pretraining} (VLP) \cite{AgrawalTN22}.

\subsubsection{Humanoid Processing: Representation, Analytics, Learning, Reasoning and Recommendation}
\label{subsubsec:knowledge}
Humanoids are associated with their data, behaviors and environments. Humanoids represent, analyze, learn, discover, reason about and recommend them to knowledge to empower humanoid functions, operations, and performance. Humanoid knowledge enables their capabilities and quality of services through analytics and learning from data, behaviors, and environments of themselves or other humanoids, robots, humans, and objects. They undertake goals and tasks to gain and strengthen their specific to comprehensive humanoid intelligence, particularly by conducting machine vision, reasoning, reinforcement learning, and learning by demonstrations, imitations, interactions, behaviors, and feedback. As illustrated on learning to communicate and interact in Sections \ref{subsubsec:connectiveness} and \ref{subsubsec:interaction}, autonomous and intelligent humanoids perform on-device and decentralized to cloud and online, shallow to deep, small to scalable, pretrained to finetuned, unsupervised to supervised, unimodal to multimodal, uni-task to multi-task, on-policy to off-policy, and specific to general learning in closed to open settings or worlds. In particular, generative AI especially LLMs and LMMs enables more scalable, self-supervised, scenario-based, and real-time humanoid intelligence, such as for real-time, interactive, affective multiparty conversation \cite{li2024ugotmeembodiedaffectivehumanrobot}, which was impossible to achieve before.

Humanoids perform knowledge representation, analytics, learning, and reasoning \cite{dstCao15,Cao24} on 1) internal data, such as the sensory information, responses, and feedback (such as failures); 2) interaction data, such as with objects, and contexts; and 3) contextual data, such as with scenarios and environmental settings. These may be self, interactive, and contextual. Humanoids perform logical, causal, probabilistic, temporal, spatial, constrained, and multimodal reasoning and inference. Further, \textit{humanoid analytics} identify, detect or predict patterns or exceptions of and insights into humanoid body parts, signals, states, behaviors, movements, and trajectories, as well as external, contextual, and environmental factors, relations, and changes. Such humanoid analytics support functional and nonfunctional objectives and different paradigms of humanoid intelligences, as shown in Fig. \ref{fig:Humanoid_robot_intelligence}, on historical, present, future/next, new/cold-start, or open/unknown data, behaviors, settings, scenarios, or contexts. 

\textit{Deep learning, LLMs and GenAI} \cite{MinRSVNSAHR24} are substantially transforming humanoids with human-like intelligence. GenAI and LLMs can empower humanoids like Ameca with unprecedented capabilities of acquiring and analyzing diverse sources, modalities, or formats of large scale inputs through pretraining and finetuning. Pretraining makes humanoid patternable or informative for signals or videos acquired from sensors of humanoid body parts; activities and behaviors of robot body or parts; interactions, communications, or coordination between robots and between robots and their environments; and transitional processes such as from vision to language or translations such as from speech to image. Humanoid pretraining builds patterns, exceptions and knowledge into humanoids, which could be further finetuned towards test, new, or open tasks or environments. Various finetuning techniques emerge and apply, such as by prompt engineering, instruction, retraining, online, test-time, contrastive, iterative, reward-based, human feedback-based, reinforcement learning, and data augmentation, to adapt pretrained humanoids to test and new tasks. 

Humanoids learn by \textit{humanoid programming}, learning by demonstrations, simulations, or imitations. Humanoids learn by demonstration \cite{ArgallCVB09}, simulation \cite{MuratoreRTYGP22}, or imitation \cite{TagliabueH24} to develop knowledge, decisions and actions from training, examples, simulation, imitation, teleoperation, shadowing, sensors on teacher, or external observations for unsupervised, semi-supervised, and few-shot supervised learning tasks. During demonstration, simulation, or imitation, domain experts or end users teach humanoids to conduct specific tasks, so that humanoids can distill knowledge or skills from these demonstrations without pre-programmed knowledge to execute new tasks. With game theory, virtual reality, and metaverse, \textit{humanoid learning by animation} will extract and discover knowledge from animated or real videos or games, and virtual/augmented/mixed reality to transfer knowledge and skills from gamers and animators, design, develop, train and evaluate novel skills and applications at a cost-effective way through humanoid-metaverse integration. 

Another humanoid learning paradigm is \textit{transformative/translational humanoid learning} (THL), where humanoids learn to transform or translate. THL  progresses from \textit{vision-language modeling} (VLM) \cite{Stone-23,KawaharazukaOKTOI24} to \textit{vision-language-to-action} (VLA) \cite{GoogleRT-2}, \textit{vision-language-to-emotion} (VLE) \cite{li2024ugotmeembodiedaffectivehumanrobot}, and \textit{vision-language-to-behavior} (VLB) modeling. These foster new directions of \textit{multimodal modeling to actions} (MMA) and \textit{multimodal modeling to decisions} (MMD) for actionable humanoid learning. In Section \ref{sec:tech-human-levelrobot}, we will further discuss humanoid analytics and directions of incorporating LLMs, GenAI and deep learning into humanoids, and specifically on opportunities including omnimodal perception-to-behavior and mind-to-action modeling for mindful, cognitive, and actionable humanoids in Section \ref{subsec:percept-beh}.

These new analytics and learning directions induce predictions, recommendations, interventions for smart humanoids. They can substantially foster and expand humanoid capabilities through simulation-to-humanoid (Sim2Real), humanoid-to-simulation (Real2Sim), animation-to-humanoid (Ani2Real), video-to-humanoid (Video2Real) transformations, and building animator-humanoid digital twin. They may further expand existing research on Sim2Real, Real2Sim, Image2Sim, and Video2Sim, etc., and tools like NVIDIA's Audio2Face.

\subsubsection{Humanoid Cybernetics: Manipulation and Control}
Humanoid cybernetics manages robot manipulation and control. Humanoids may manipulate their body parts such as eye movements, facial expression, or postures, or external objects such as grasping objects.  
\textit{Humanoid manipulation} \cite{murphy2019introduction} serves various tasking purposes and involves diversified robot parts, which may be classified into typical types including system-centric manipulation, object-centric manipulation \cite{Stone-23}, action or state-centric manipulation, transition-centric manipulation, and process-centric manipulation. System-centric humanoid manipulation operates a robot part, e.g., manipulating the locomotion. Object-centric humanoid manipulation instructs, controls and manipulates robots toward specific objects under object background. Action-centric and state-centric humanoid manipulation guides, controls and operates robots towards taking specific actions or approaching target states, e.g., rearranging an object or reorienting a robot hand. Transition-centric humanoid manipulation focuses on robot state or behavior transition from one to another, e.g., raising a hand and then positing it toward a target object. Process-centric humanoid manipulation involves multiple steps of action or state progression or change, e.g., from perceiving the environment to moving a hand and then grasping an object.

\textit{Humanoid control} represents a strong type of manipulation of humanoids or external objects, where a humanoid influences, directs or manages its behaviors, actions or their processes.  
The objectives and tasks of humanoid control are varied \cite{ProiaCCD22}. Humanoids may control its parts such as head, eyes, emotion, gait, legs and hands. They may perform control to implement their functions such as trajectory, horizon, gesture, posture, collaboration, interaction, behavior position, and speed. Humanoid control can execute specific tasks such as bipedal locomotion, omnidirectional walking or satisfy nonfunctional requirements such as safety. Humanoid robots may be controlled by different mechanisms. Functionally, humanoid control relies on the corresponding mechanical and electrical humanoid designs to implement specific objectives and tasks, such as optimization methods, dynamic models, bionic methods, reinforcement learning, demonstration learning, imitation learning, and prediction. Controls are incorporated into humanoid body parts, such as by a central nervous system, a motor controller, stability and balance measures such as zero moment point, capture point and centre of pressure, and data- and model-driven analytical and learning techniques. Control environment may be open, closed, semi-open or semi-closed in terms of interactions with environments. Control modes may be autonomous, semi-autonomous or hand-engineered in a static, dynamic or adaptive manner. Automated control may be driven by policies through real-time learning or rules predefined or hand-coded. 

\textit{Learning-to-control} and data-driven and behavior-based control represent new control modes and scenarios, where analytical and learning systems such as anomaly detection, predictive learning and generative AI are incorporated into robot engine for monitoring, detecting, predicting and preventing control failures, damages or accidents \cite{SubburamanKTL23,WangZ24}. In AI-empowered humanoids, a typical technique of learning-to-control is deep reinforcement learning, which obtains optimal actions or state-action combinations, i.e., policies and strategies, to decide and govern humanoid behaviors and actions. Humanoid learning-to-control may also apply learning by demonstration, imitation, prediction, and recommendation etc techniques to obtain control capabilities.  

Humanoid robots also involve new control problems or approaches, which may relate to humanized features, functions and tasks. For instance, facial expression control, eye and mouth movement control, human gesture and posture control, hand and leg movement control, and human-humanoid interaction can be guided by hand-engineered rules or policies or learning-driven findings. These require not only ergonomic design but also control, in particular, real-time, interactive and online control from multitask, multimodal and omnimodal VLA modeling and cognition-to-action translation from web-scale data, humanoid data, and external environments.

\subsubsection{Humanoid Generalization: Adaptation and Optimization}
Humanoid generalization enforces adaptation and optimization. 
Humanoid adaptation adjusts the states, behaviors/actions, policies/strategies, or even parameter settings of a humanoid or a humanoid team to fit their evolving goals, plans, tasks, processes, conditions, contexts or environments. Adaptation applies to all humanoid tasks. Such humanoids should have the adaptability and corresponding functions to sense changes, feed back deviations, arrange strategies, adjust actions, evaluate performance, and reflect on adaptation. Adaptive humanoid designs and capabilities include mechanical design with adaptive sensor motors, adaptive control techniques, online and active perception and recognition, evolving human-robot interaction, evolving learning, dynamic finetuning, predictive control, online motion generation \cite{MengYCHMH24}, predictive next actions, and proactive risk and compliance management. 

Specifically, \textit{general-purpose humanoids} may live in real-life environments and face dynamics and changes of states, functions, and tasks in semi-open to open worlds, e.g., during conducting real-time automated capabilities. Accordingly, how to make general-purpose humanoids to act, reason and interact like a human being in real-world scenarios poses a fundamental challenge to humanoid robotics. Existing approaches such as sharing \textit{mental models} across all robots and enabling predefined flexibility for bipedal locomotion and dexterous manipulation in known evolving environments cannot capture novel or unseen scenarios. General-purpose humanoids in open, novel and uncertain dynamics require \textit{learning to adapt}, or \textit{adaptation by learning}, and human-supervised adaptation \cite{WonsickLOWP21}. Humanoid adaptation by learning may demand predictive architectures, data-driven adaptive learning capabilities, neuro-inspired control, and evolutionary learning. For example, adaptive evolutionary humanoids may amalgamate biomimetic design, morphological computation, and evolutionary learning to enable adaptive mutation, crossover, selection, and optimization of actions, states, policies, and goals iteratively and sequentially. 

Humane real-world scenarios demand a real-time understanding and transferring human intention, emotions, empathy, and decision sequentially to humanoid tasking over time. In a humanoid or human-humanoid society, humanoid adaptation further requires understanding and obeying social norms during social interactions and activities.

\subsubsection{Humanoid Design: Multiaspect Integration}
Anthropomorphic and aesthetic humanoids are driven by their mechanical, electronic, electrical, biological, and social designs concerning robotic hardware, software and operational developments. These humanoid designs surpass robot designs  \cite{SaeedvandJAB19} to empower humanoid-oriented sensors, processing units, mechanical structures, kinematics, electronic and electrical devices, power supply, and software architectures and implementations incorporated with humanoid social norms and activities to imitate human-like and lifelike appearance, functions, behaviors, and intelligence. 

\textit{Humanoid mechanical design} involves new technologies \cite{Ficht21,Hashimoto20} to enable anthropomorphic body and parts, such as android head, neck, torso, hand, fingers, leg, and foot, their covering skin and muscles, and their proportions, and distributions; skeletal and musculoskeletal structures with joint mechanisms with actuators and drive mechanisms for different body parts such as for hip, knee, and ankle; android operations and control, such as locomotion mechanisms, force and strength manipulation, legged or wheeled systems, joint movement, and transduction mechanisms and functions; anthropomorphic functions such as eye gaze, touch sense, grasping intensity, and assembly; humanoid behaviors such as mechanical enablers for movement, navigation, and tracking; android aesthetics for enabling human-looking features such as eyeballs, eyebrows, and lip \cite{Ishiguro2024d}; and humanoid control and assurance such as enabling humanoid safety, and preventing fall or slippage. 

\textit{Humanoid electrical design} integrates electrical and electronic technologies into robotic and humanoid structures, functionality, reliability, and adaptability, etc. \cite{JAS-Tong23}. Electrical designs enable humanoid sensors for proprioception, exteroception and processing; electrical actuators and motor drivers; power systems; networking and communications; embedded systems and integrated units; on-board signal and information processing; human-humanoid interaction systems; and quality assurance such as safety and failure management. These sensory and electronic mechanisms and functions are integrated with mechanical designs, mechanisms, and functions, as well as other mechanisms and systems such as resistive, capacitive, optical, piezoelectric, and magnetic mechanisms and devices. For humanoids with rich anthropomorphic features, such as the realistic Ameca head \cite{Ameca}, advanced motors and articulation systems under the silicon skin drive and control nuanced lifelike features, functions, and performance, such as facial muscles, expressions, emotions, eyebrow movements, nose crinkle, lip and jaw movements, and arm movement. 

\textit{Humanoid biological design} is bioinspired and bionic. Biological design simulates biological and biomedical mechanisms, structures, materials and control of the human body in humanoids to create more natural, human-like, lifelike and rational structures and functions and more intricate tasking and performance of humanoids \cite{LahrYH16,Pfeifer07}. Biological design specifies the appearance, structures, functions, behaviors, and performance of humanoid body and parts by mimicking various biomedical, ergonomic and neural mechanisms, parameters, and settings. They specify artificial body of humanoids \cite{Blackiston22,Ishiguro2024d}, including skeletal framework, artificial skin, muscles and tendons; synthetic textures, responses and degrees of freedom; physiological states and conditions, such as size, weight, strength, load, volume, mass; and neuromorphic features, such as emotional expression and mental states. Biological design inspires the definition and implementation of humanoid functions, such as walking, climbing, vision, hearing, and touch; and humanoid behaviors such as gestures, postures, mobility, interaction, and learning. 

\textit{Humanoid social design} defines social (interactive, collective), ethical, and emotional/affective mechanisms, activities, and norms of humanoids in a humanoid team or society or with humans \cite{Korn2019}. Such social design provides technological supports (both connecting to mechanical, electrical and biological designs and sociality-specific designs \cite{KhavasAR20}), social and ethical specifications, and even affective engagement for humanoids to interact, collaborate, cooperate, team, compete or even attack/defend with humans or other humanoids or agents individually or collectively. Human-humanoid interactions, teaming, collaborations or competition require enforcing social, ethical, emotional, psychological and cultural rules, norms or checklists for dos and don'ts in relation to their contracted goals and tasks, interactive objects and attributes, and embodied or embedded physical, remote or virtual situations, scenarios and environments \cite{FongND03}, such as avoiding collision, and prohibiting attacking humans. Differing from non-humanoid social robots, social humanoids may require certain levels of human-likeness, life-likeness, humanness, and even humanity to facilitate humane interactions and address the uncanny valley effect. Social design also involves nonfunctional matters as discussed in Section \ref{subsec:nonfunctional}, including \textit{humanoid ethics} to regulate social humanoids by trust, safety and privacy checks before and during interactions and enforcing their responsibility, and accountability. 

These humanoid mechanical, electronic, electrical, biological and social designs guide the implementation and integration of humanoid structures, senors, actuators, communications, interaction, networking, interface, and system performance.

\subsection{Nonfunctional Specifications of Humanoids}
\label{subsec:nonfunctional}
With the fast growth and advancement of robotic techniques and functions, humanoid robots are expected safe and secure in their operations and action-taking; trustful and responsible for their actions; transparent and explainable in their outputs; empathetic and rational during teaming, collaboration and interaction; and compliant, legal and ethical for integrity and regulation.
Accordingly, the functions and quality of humanoids need to be ensured by various nonfunctional requirements, objectives and constraints. As summarized in Table 3 \cite{HumanoidAI25}, they cover many aspects from traditional requirements on the robotic quality of services to  new control problems relating to humanized requirements. We categorize such nonfunctional specifications into four aspects below: 1) human likeness satisfaction, 2) humanoid capability maturity, 3) humanoid performance measurement, and 4) humanoid impact estimation.

\subsubsection{Human Likeness Satisfaction}
\textit{Human likeness satisfaction} refers to the extent a humanoid resembles the cognition, appearance, behaviors, biometrics and social interactions of humans and their lives and the cultural alignment with them. These human anthropomorphism and life likeness of a humanoid can be categorized in terms of the humanoid cognitive likeness, physical likeness, biological likeness, behavioral likeness, and sociocultural likeness of humanoid design and development.

\textbf{Humanoid cognitive likeness} refers to the extent that a humanoid resembles the cognitive, mental and subjective mechanisms, processes, states, traits, and abilities. It replicates human thought process, logic, intentionality, intuition as well as characters and personality \cite{Tabrez2020ASO}. Cognitive likeness also measures the humanoid ability to think, understand, communicate, analyze, learn, reason, infer, generalize, respond, and adapt as a human. In humanizing and personifying robots toward human-like, human-level and humane intelligent systems, humanoids are further humanized with mechanisms, properties and traits such as human-like aggressiveness, morality, consciousness and personality \cite{LuoOPI22}. For example, a humanoid may present extroversive, agreeable, conscientious, neuroticive, open or aggressive personality, forming \textit{humanoid personality} in articulating robotic features and functions and in robotic tasking and employment. A humanoid may be rational with intentionality and consciousness for being or doing good, with moral and empathetic actions and behaviors during tasking or human-humanoid interaction, teaming and collaboration, which creates \textit{humanoid intentionality} and \textit{humanoid rationality}. Such humanoid personality, intentionality and empathy require bidirectional design and support. On one hand, a humanoid can access a human or other robot's cognitive, emotional and affective states, and then behaves and interacts with them with empathy and morality and meeting human comfort. On the other hand, humans can fully access and control a humanoid's intentional stance, and personality, and their behaviors can be managed and interpreted by a human being.

\textbf{Humanoid physical likeness} measures the extent a humanoid resembles human physical structures, attributes, functions, and aesthetics by visual appearances, layouts, and interfaces \cite{Asano17}. These may be measured in terms of their resemblance dimensions, proportion, ratio, intensity and likelihood of a human body structure, facial components and features, and skin and surface texture. For example, Ameca Desktop focuses on mimicking human facial likeness, while the full-body Ameca \cite{Ameca} expands the anthropomorphism and aesthetics to the entire body of the humanoid. Cybernetic avatars \cite{Ishiguro2024d} present nuanced human appearance including artificial skin, eyeballs and eyebrows. 

\textbf{Humanoid biological likeness} assesses the extent a humanoid replicates the internal and external systems, functions and features of human biological systems. Biological likeness complements physical likeness with the underlying bioinspired materials, structures, processes, behaviors and features of human biology and physiology \cite{Blackiston22,LahrYH16}. The resemblance of human sight, tactile, olfactory, taste, and hearing contributes to the phenomenal sensory capabilities, design and mechanisms in humanoids. In addition, a humanoid may mimic the skeletal framework of human anatomy for structural integrity and joint designs; the features, abilities, functions, sensitivity and adaptability of human skin and muscles for surface coverings, movement and dexterity. A humanoid may even resemble human biomedical features, e.g., nervous systems for sensory input and motor control, circulatory systems for energy and movement circulation, and respiratory and energy systems for metabolism, storage and control. 

\textbf{Humanoid behavioral likeness} refers to the extent that a humanoid emulates the actions, reactions, emotional expressions and interactions of human body parts and systems. Recalling humanoid behaviors in Section \ref{subsubsec:activeness}, behavioral likeness captures the body language, gestures, postures, and movement of body parts of humanoids over humanoid physical likeness. These reflect the design and functionality of humanoid locomotion in reproducing human-like movement mechanisms, styles and patterns \cite{MengYCHMH24}, such as for humanoid walking, grasping, and dynamic adaptation and balance. Further, the likeness of humanoid gait, facial expressions, tone of speech, spoken and motion emotions facilitates human-like verbal and nonverbal communications, affections and interactions. In social contexts \cite{Cao10}, behavioral likeness enables human-like social behaviors, teamwork, cooperation and collaboration, and adaptive responses during social interactions between humanoids and between humanoids and humans.

\textbf{Humanoid sociocultural likeness} refers to the extent that a humanoid exhibits its functions, attributes and behaviors aligned with human ethical, moral and social norms, cultural appropriateness and values, and sociocultural traditions and boundaries during social contexts and dynamics. Social and cultural likeness characterizes humanoid abilities of respecting social roles, cultural cues, sociocultural norms and religious sensitivity while undertaking sociocultural tasks, behaviors, interactions and engagement \cite{Korn2019,FongND03}. It determines the extent that situational humanoids are embodied with culturally relevant appearances, symbols, and settings and have the awareness of social and cultural norms, courtesy and etiquette. Humanoids demonstrate the language and dialect adaptation in human-humanoid conversations, the culturally appropriate gesture and body language during nonverbal communications, and the culturally sensitive emotional expressions and responses. Sociocultural humanoids also understand, respect and exhibit social intelligence, commonsense, empathy, and emotional appeal for engaging and pleasing user experience, and situational reasoning during social activities and contexts.

\subsubsection{Humanoid Capability Maturity}
\label{subsubsec:capabilities}
\textit{Humanoid capability maturity} measures the proficiency of ability, functionality and sophistication of humanoids in pursuing their objectives and performing their tasks. Humanoid capabilities traverse cognitive, perceptual, physical, behavioral, emotional, communicational, linguistic, and interactive abilities. We structure humanoid capability maturity into three dimensions: \textit{intelligence capacity} of reproducing human intelligence, \textit{functional capability} of undertaking tasks, and \textit{complexity resilience} in handling sophisticated tasks, data and environments. While these may involve nonfunctional capabilities of humanoids, here we focus on functional competence which may be quantified in terms of their maturity levels and assessment specifications, criteria and measures. 

\textbf{Humanoid intelligence capacity} assesses the spectrum,  capabilities and intensity of intelligence paradigms of humanoids in replicating human intelligence. Intelligence capacity spans over and meta-synthesizes diverse qualitative to quantitative human intelligence \cite{aikp_Cao15}, including \textit{cognitive intelligence} for humanoid mental mechanisms, intentionality, and personality; \textit{perceptual intelligence} of humanoid vision, audition, and other sensory mechanisms and abilities; \textit{emotional intelligence} of humanoid emotion recognition and simulation; \textit{interaction intelligence} of humanoid collaboration and competition, and verbal and nonverbal communication; \textit{social intelligence} of humanoid-human interactions, cultural adaptation; \textit{data intelligence} of humanoid data acquisition, analytics and management; and \textit{learning intelligence} of humanoid representation, learning, reasoning, memorization, and generation. The capacity of humanoid intelligence builds on AI techniques and approaches \cite{Cao22b} for implementing the paradigms of human intelligence. A humanoid may be empowered by individual or hybrid intelligences to perform their tasks and fulfill their functions. 

\textbf{Humanoid functional proficiency} refers to the degree of competence and skills of a humanoid in performing their tasks and functions illustrated in Section \ref{subsec:functional}. Humanoid functional proficiency can be customized into \textit{cognitive proficiency} of implementing humanoid brain, \textit{physical proficiency} of implementing anthropomorphism, \textit{perceptive proficiency} of performing vision, speech, hearing and other sensory functions, \textit{behavioral proficiency} of achieving activeness, \textit{connective proficiency} of performing networking and communications, \textit{interactive proficiency} of conducting collaboration and competition, \textit{organizational proficiency} of planning and search, \textit{knowledge proficiency} of representation, analytics, learning, reasoning and recommendation, \textit{cybernetic proficiency} of conducting manipulation and control, \textit{generalizable proficiency} of achieving adaptation and optimization, and \textit{system proficiency} of humanoid design and implementation. Humanoid proficiency may also be categorized into \textit{task-specific proficiency} in undertaking a specific task, \textit{nonfunctional proficiency} and other specific types of proficiency \cite{NortonACFGGSSSS22}, such as \textit{emotional proficiency} of demonstrating empathy, and \textit{social proficiency} of engaging humans and adapting to social, ethical and cultural norms. Most humanoids as shown in Table 2 \cite{HumanoidAI25} only possess basic to intermediate levels of proficiency of some of these aspects. For example, few humanoids like Ameca have the potential of incorporating \textit{cognitive, sentimental or emotional states and features}, such as facial expressions. In addition, social humanoids may hold humanized features and traits, such as personality as embodied by extroversion, agreeableness, conscientiousness, neuroticism and openness \cite{FongND03}. 

\textbf{Humanoid complexity resilience} refers to the ability of humanoids to robustly, flexibly and adaptively perform their tasks and functions in uncertain, sophisticated, difficult, and unexpected systems, conditions and environments \cite{dstCao15}. Complexity resilience measures \textit{goal complexity resilience} of complex humanoid objectives and expected outcomes; \textit{task complexity resilience} of humanoid task complexities such as complex task settings, dynamics and evolution; \textit{system complexity resilience} of humanoid design and implementation complexities; \textit{data complexity resilience} of humanoid data complexities such as heterogeneity, non-IIDnesses and interactions \cite{Cao14}; and \textit{environmental complexity resilience} of environmental and contextual complexities such as heterogeneous, dynamic and coupled contexts. Humanoid resilience can also capture other aspects, such as \textit{cognitive resilience} in handling cognitive goals, tasks and contexts; \textit{physical resilience} in building robust structures and components and efficient designs; and \textit{social resilience} to cross-cultural contexts and diverse social practices. Humanoids are still in the early stage of embedding complexity resilience, with the current focus on task robustness in multimodal and changing contexts \cite{TagliabueH24}.

\subsubsection{Humanoid Performance Evaluation}
\label{subsubsec:performance}
\textit{Humanoid evaluation} verifies and validates functional and nonfunctional performance of executing their requirements and specifications. The evaluation criteria and measures are both objective and subjective, and technical and nontechnical. The aspects to evaluate spans over humanoid objectives, designs and performance \cite{KandaIIO04}, from state to behavior, from domestic to social, from individual to collective, from physical to virtual, and from implicit to explicit. These evaluation aspects can be inspected per \textit{task performance} at design time, \textit{operational performance} at running time, and \textit{application usability} by end users and for applications through design evaluation, simulation testing, field testing, stress testing, and user feedback. These evaluate whether a humanoid is technically and operationally capable, effective, efficient, usable, compatible, portable, trustful, responsible, explainable, transparent, safe, compliant, secure, accountable, reproducible, maintainable, and interoperable.  

\textbf{Humanoid task performance} evaluates how good a humanoid conducts its designs and functions, i.e., the quality of functions. Functional performance can be assessed in terms of goodness of functionality, design and tasking \cite{dstCao15,SaeedvandJAB19}. 
First, the \textit{goodness of humanoid functionality} assesses the quality of functions, i.e., how good a humanoid performs its functions from part to whole. Functionality goodness can be quantified in terms of the completeness, creativity, maturity, effectiveness, stability, robustness, versatility, agility, dexterity,  objectivity, trustfulness and responsibility of humanoid functions and technical capabilities. Humanoid robustness evaluates the generalizability of humanoid functions and designs. Humanoid objectivity measures the fairness and unbiasedness of humanoid components, algorithms, and their decision-making and results. Humanoid trustfulness evaluates the unidirectional, reciprocal and bidirectional human-humanoid reliability and confidence by humans in a humanoid's roles, capabilities, actions, functions, decisions, and performance. 
Second, the \textit{goodness of humanoid tasking} examines the quality of performing tasks by customizing and configuring functions for specific tasks, scenarios, and use cases. Tasking goodness can be evaluated in terms of flexibility, configurability, adaptability, comparability, transferability and capacity of applying functions to ad hoc, specific, multiple and domain-oriented tasks and task execution performance. Task execution may be estimated by quantitative to qualitative performance measures, such as requirement met rate; throughput and capacity of transactions, tokens, requests and users to accommodate; successful completion rate, and task success rate; consistency within tasks and comparability across tasks and scenarios; adaptation to different tasks, domains, contexts, and applications, as well as common measures like accuracy and precision. 
Last but not least, the \textit{goodness of design} measures the quality of design. Design goodness may be evaluated in terms of measures such as design modularity, analyzability, reliability, scalability, transparency, explainablity, accountability, testability, reusability, reproducibility, and recoverability.
Examples include modularity of interchangeable and independent components; reliability of available operational uptime, error and fault tolerance, failure rate, mean time between failures; durability of functionality and performance over time, changes or failures; analyzability of understanding algorithm, system and decision and diagnosing problems; scalability over user, load and request increase; reusability of components; recoverability from failures or faults and recovery time and effort; and testability of functional modules and their integration.  

\textit{Humanoid trustfulness and responsibility}.  
Factors influence human trust in humanoids may be various \cite{KhavasAR20}, e.g., lack of emotional intelligence, limited ability to adapt to new situations, unclear decision-making processes, privacy concerns, negative media portrayal, unclear code of ethics, or regulatory oversight. The trustfulness and responsibilities of humanoids are embodied by various functional and nonfunctional mechanisms and measures, such as reliability, adaptability, security measures, programmed ethical guidelines, intention transparency, ethical behaviors, ability to self-control and self-regulation, trustful past track record, and subjectivity such as empathy and responsibility. These humanoid trust factors and measures are embodied by the trustfulness \cite{ChoCA15} in humanoid design, behaviors, effect and performance, such as automation, live recognition, and real-time prediction and decision-making; the predictability and transparency of risk, safety, security and consequence; the detection, prevention and resolution of system issues such as faults, unsuccessful task completions, and irreparable mistakes.
As subjective properties increasingly affect the humanization of robots, enforcing the morality of humanoids essentially enables what moral liability, accountability, rights and responsibilities a humanoid should hold and how such attributes should be distributed in multi-party settings. However, humanoid empathy would not serve as the final guard rail. Humanoid trust may further require robotic legal rights, such that a humanoid must not incur human injuries and must obey orders. This triggers the study on the rights of humanoids and artificial consciousness, e.g., whether a humanoid is a legal entity, and to what extent a humanoid can be given what legal rights comparable to humans. Humanoid legal rights are crucial for humanoid services like law enforcement and being a robo-adviser, an elderly companion, or a child caregiver.

\textbf{Humanoid operational performance} evaluates how good a humanoid is operated, i.e., the quality of operations (or quality of services). Operational performance can be evaluated against performance measures \cite{dstCao15}, such as autonomy, standardizability, operability, interoperability, compatibility, transparency, explainability, efficiency, elasticity, safety, privacy preservation, risk compliance, and cost-effectiveness. 
\textit{Humanoid autonomy} refers to the level of automation when a humanoid operates and governs independently. 
\textit{Humanoid standardizability} measures the adherence to widely-acceptable development, industry and data exchange standards or protocols. 
\textit{Humanoid operability} measures operational competence under different scenarios and conditions, such as online and cloud-based development support, deployment environment of terrain and temperature,  and service and testing facilities. 
\textit{Humanoid interoperability} and \textit{compatibility} determine the connection, communication, interaction, collaboration, and integration of a humanoid with other robots, devices, modules, systems, and environments, and mobility in different terrains and environments. 
\textit{Humanoid efficiency} may be quantified by measures such as computational efficiency, latency and response time of task and request, task execution time, CPU and GPU time, memory cost, bandwidth, resource consumption, energy efficiency, and decision-making efficiency.
\textit{Humanoid elasticity} measures performance stability under changes, such as humanoid workload changes without efficiency degradation, and consistency across tasks.
\textit{Humanoid adaptability and flexibility} measure operational performance under new or unexpected conditions, environments or contexts, with new objects or tasks,  over time or space, and with responses to dynamics, changes and unexpected scenarios. 

The \textit{humanoid transparency}, or opacity, of robotic designs, behaviors and outcomes affects human confidence in humanoids and their applications and services. Two mechanisms to ensure humanoid transparency are explainability and reproducibility \cite{gunes2022reproducibility}. \textit{Explainability} provides mechanisms, tools and interfaces to understand and interpret humanoid designs, working mechanisms, behaviors, outputs, and consequences. \textit{Reproducibility} ensures the repeatability, consistency and integrity of humanoid designs and outputs under the same circumstances. Explainability and reproducibility become increasingly essential for complex humanoid designs and tasking. Transparency, accountability, auditability, explainability and reproducibility are essential in implementing humanoid automation, and on-the-fly and in-the-wild design, operation and decision-making with open settings. 

\textit{Humanoid safety}. \textit{Physical safety} includes no unintentional or unwanted contact between humanoids and humans or comfortable physical contact with force below thresholds or without physical harm \cite{LasotaFS17,ProiaCCD22}. \textit{Psychological safety} comprises indirect psychological harm, discomfort or stress such as human trust, unexpected intention, and negative emotions. Safety can be ensured by careful programming, safety criteria and constraints such as biomechanical limits, injury prevention, control strategies, and assurance mechanisms such as detecting, predicting, preventing and intervening with force, collision, falling, crash, risky and abnormal activities. In open settings, safety monitoring, detection, prediction, prevention and intervention are essential for novel, cold-start, evolving and changing scenarios and accidents. \textit{Humanoid security} further enforces requirements on data and programming integrity with authentication and modification restriction, confidentiality of resources and IP for authorized users, authorization and permissions for users and developers, and auditing by monitoring, logging and tracing actions and non-repudiation of actions.

\textit{Humanoid risk and compliance}. A humanoid may be exposed to various risky scenarios or actual risks \cite{jeeva2023risk}. They include methodological and technical risks, such as fragile generalization,  faulty design, and incomplete compliance; malicious manipulation such as misuses, misinformation, malicious attacks, and manipulated behaviors and automation; and ethical issues such as deception and privacy concerns. Humanoids may also exhibit unexpectedness such as hallucination of automated generation, and singularity and superintelligence of self-improvement and intelligence emergence, when they gain control and rights beyond human perception and acceptance. 
As automating real-time, interactive, decentralized and personalized humanoid tasks and operations become predominant, \textit{humanoid surveillance} not only requires automated, real-time, tailored and predictive risk and compliance monitoring, diagnosis and management, but also human surveillance i.e. keeping human in the compliance loop. In this regard, significant challenges exist in real-time and automated scenario-oriented human-humanoid regulation, e.g., in accountable human-humanoid role allocation and responsibility sharing, efficient regulation in complex settings, privacy-preserving surveillance, risk-aware human-humanoid cooperation, and essential human-centered control and mitigation.

\textbf{Humanoid application usability} assesses the applicability and user-friendliness of a humanoid in applications. Evaluation criteria and measures may include humanoid usability, deployability, installability, maintainability, satisfaction, learnability, accessibility, and recoverability \cite{Ishiguro2024d,casiddu2021humanoid}.  
\textit{Humanoid usability} can be assessed by interface and interaction friendliness; ease of operations and use at different platforms, channels, network conditions; user experience, acceptance and satisfaction; learnability by new users; accessibility of disabled, aged or disadvantaged users; use efficiency of time and effort; error tolerance to errors, mistakes; noise level and disturbance; and ease of recovery. 
\textit{Deployability} measures humanoid conditions and constraints of deployable applications, areas, environments, facilities, networking, communication, and space.
\textit{Installability} measures how easy to install, configure, debug and test the humanoid, its plug and play, and conditions and requirements on installation technicians and services.   
\textit{Maintainability} assesses the competence of modular design principles; ease of repairs, updates, modifications, and upgrades; recovery difficulty level; and maintenance interfaces, guidelines and wizards.
\textit{Supportability} evaluates the technical support e.g. by documentation, visualization, virtual reality or training, and their service costs and time availability.

\subsubsection{Humanoid Impact Estimation}
\label{subsubsec:impact}
The outcomes of applying a humanoid foster humanoid impact across intellectual, economic, environmental, social and ethical aspects. They measure the impact and benefits of humanoid technical advancement, market valuation, environmental sustainability, societal contribution, and ethical implication.  

\textbf{Humanoid intellectual impact} measures the influence of humanoid technologies, systems and processes on technological innovation, knowledge advancement and creation, and problem-solving capabilities. 
Humanoid technology fosters a new research area towards technological innovation and creation, and new research and development opportunities. This spans over both foundations of enabling and developing humanoids and applications of humanoids in diverse domains and sectors. The various and significant humanoid intellectual impacts include driving technical competitiveness, intelligence level, and automation; fostering new humanoid-related knowledge advancements, expansion and accessibility; augmenting human autonomy of amplifying human intellectual capabilities, efficiency, and performance; enabling human-humanoid collaborative superintelligent systems and services, unresolvable by human alone; amplifying personalized learning capabilities, methods, experience and effect; cultivating interdisciplinary innovation by integrating humanoids into other fields such as humanoid health and medicine; and inducing skill development and upgrading, and expertise transfer. These intellectual impacts will further drive economic, social, ethical and environmental impacts, such as promoting industrial performance improvements, cognitive load reduction, and production efficiency.
Intellectual impact can be measured by relevant factors, such as technological advancement level, creativity level, technological adaptability, task efficiency,
task success rate, and problem-solving efficiency.

\textbf{Humanoid economic impact} is embodied through humanoid-enabled and -augmented economic opportunities and benefits \cite{ruiz2024impact,dstCao15}, such as novel products and services, job creation, productivity gain, efficiency augmentation, cost reduction, market expansion, industry transformation, and economic resilience.  
Humanoid technology is boosting the \textit{humanoid industry} and  \textit{humanoid economy}, fostering new humanoid-driven industries, products, services, and activities \cite{CaoDeAI}. The deployment of humanoids to diverse business domains, industry sectors, and social areas promotes \textit{productivity gain} and \textit{efficiency augmentation} such as by new manufacturing industry, industrial automation, augmenting production efficiency, automating production and services, and improving service quality and delivery by humanoids. Humanoid technology also transforms existing industry and economy, in particular, education, services, marketing, entertainment, and hospitality \cite{dstCao15}. Humanoid can also deliver unique economic and social impact in areas like disaster response and rescue, agriculture, and defence and security industry.
The \textit{humanoid ecosystem} further drives job creation, displacement, up-skilling and transformation with new job roles and responsibilities from humanoid R\&D to production, logistics, sales, marketing and technical services. 
The economic resilience can be strengthened by humanoid technology, including humanoid technological affordability, maintainability, extendability, and upgradability of their purchase, operations and maintenance in daily work and life. Humanoid technology produces advantageous financial and productivity outcomes, measurable by economic impact factors, such as competitive advantages, market acceptance, automation level, production efficiency, revenue growth rate, cost efficiency and cost-benefit ratio, productivity gain, reduced operational costs, enhanced personalization of intelligent services.

\textbf{Humanoid environmental impact} refers to the influence of humanoid technology on sustainable development and environmental areas, benefits and opportunities. The environmental impact can include developing sustainable humanoid technology, industrialization and economy, and contributing to environmental sustainability. 
\textit{Sustainable humanoids} comprise computing-, resource- and energy-efficient, sustainable humanoid design, innovation, development, production, logistics, services, and practices. Examples include light-weighted architectures; restricted consumption of CPU, GPU, network, communication and energy; resource utilization in production and operation; technological sustainability such as eco-friendliness and net-zero design, durability, maintenance costs, and lifecycle assessment; longevity such as period of functioning, expiry period, cost of maintenance, and extendability over time; and modular compatibility with other technologies and products.
\textit{Humanoid environmental sustainability} measures the ecological consequences of humanoid technology and eco-friendly and sustainable applications and practices. Examples of humanoid ecological outcomes include depletion, emission, waster and pollution generation and reduction by humanoids; and recyclability of humanoid materials, components, and design. Examples of sustainable humanoid applications include protecting and restoring damaged environments; enabling renewable energy integration; and monitoring, analyzing and intervening environmental changes. Humanoid applications to environmental areas include humanoids for precision farming, weeding, irrigation and pest control; smart urban, home, society, city, transport, waste, recylcing and utility management and monitoring; diaster response; and wildlife protection and conservation.

\textbf{Humanoid social impact} refers to the influence of humanoid technologies on human lives and societal areas, benefits and opportunities. Social impact may include improving social businesses and applications, social diversity, equity and inclusion (DEI), social interactions and community building, public safety and security, technology democratization, and life quality and social well-being through a societal acceptance of humanoids. 
Humanoids can benefit \textit{social businesses} \cite{Korn2019} such as healthcare, welfare and wellness through assisting in urban-based, aged, elderly, disabled, or marginalized individuals and communities. Humanoids can promote \textit{DEI objectives} such as by expanding education accessibility, knowledge transformation, and personalized services to those disadvantaged areas and populations. Humanoids can also facilitate \textit{social connections}, cultural engagement and community cohesion. The roles of humanoids in monitoring, analyzing, intervening and improving \textit{safety, emergency, threat, hazard and disaster} accidents and aiding in search and rescue responses. Humanoids have been used to improve \textit{life quality}, such as engaging transportation, animation, digital arts, entertainment, and hospitality \cite{FongND03}. Humanoids can support \textit{democratization of technologies and knowledge} towards improving individual, organizational and social capabilities and benefits in rural, disadvantaged areas and less technology-savvy domains and communities; and \textit{equity} between social groups, such as equitable access to humanoid benefits by bridging digital and technological barriers.

\textbf{Humanoid ethical impact} refers to the influence of humanoids on ethical concerns and opportunities. \textit{Ethical concerns} traverse uncanny valley effect, disruptions, and humanoid transparency, trust, fairness, privacy, vulnerability, and accountability \cite{Dumouche23}. \textit{Ethical opportunities} include addressing misinformation, improving humanoid humanization and intelligence, and leveraging human intelligence.
Addressing eerie feelings of humanoids or negative emotional response of humans by human-looking humanoids concerns about the humanoid humanity dilemma - a humanoid realization of ``almost but not quite" human, and the loss of human uniqueness and identity. Addressing this effect for widespread societal acceptance of humanoids should accommodate humanity into humanoid anthropomorphic likeness and address technical, material and operational complexities and constraints.
Humanoids may induce social, cultural, societal, economic and ethical disruptions such as job displacement and humanoid-human-coexisting societies and norms. Humanoid ethical and technological adaptation must articulate subjective human traits in humanoids, such as humanoid personality, and societal, moral and legal considerations and requirements \cite{Korn2019} on humanoid responsibility, trust and ethics. The development, management, governance and regulation of humanoids and their data, algorithms and platforms also require corresponding ethical, legal norms and laws, such as enforcing traceability to specific metrics, factors and modules; accountability of ownership and responsibility; transparency of decision criteria, process and outcomes; privacy concern in data processing, collection, algorithmic recommendation, and parameter sharing; and safety concern such as on human incidents, compliance with safety standards and design specifications, and threat or trouble to users and surroundings. 
While pursuing human-centric humanoid attributes, aesthetics and automation, it is also important to balance human creativity and humanoid intelligence, and assure humanity, human identity and well-being, including enforcing risk compliance on potential disinformation, misinformation, and fake information.

\textit{Humanoid ethics}
Robot ethics
\cite{Dumouche23} concerns about the privacy, manipulation, opacity and bias of robotic systems and the equilibrium effect and consequence of robotic designs, mechanisms (e.g., autonomy and predictive decision-making) and behaviors during operations, decision-making and human-robot interaction. \textit{Ethical humanoids} are built with transparent, fair and unbiased design and decision-making; proactive compliance, due process and auditing to potential effects, influences, vulnerability, deception, misinformation, synthetic issues, failures, changes and uncertainty; and accountable compliance, interference and intervention measures and mechanisms. Humanoids may further involve humane aspects such as confirmation bias, deception, malicious design, attack and manipulation. Depending on the level of ethical enforcement, a humanoid may be explicitly ethical, implicitly ethical, or fully ethical.

\section{Realistic Humanoids: Other Perspectives}
\label{sec:tech-human-levelrobot}

The comprehensive analysis of existing humanoids in Section \ref{sec:humanoidfamily}, as outlined in Table 2 \cite{HumanoidAI25}, and the systematic discussion on functional and nonfunctional specifications of humanoids in Section \ref{sec:requirements} disclose various gaps, limitations of existing work and the implications for building new-generation humane humanoids. Fig. \ref{fig:hrobot-family} highlights other perspectives and directions in developing humanoid AI: humanoid humanity, mind-to-action, biomedicine, animation, virtual-real humanoid mapping and symbiosis, metaverse, humanoid-human symbiosis, generation, decentralization, standardization, and quality of services. 
With space limitation, this section expands Section \ref{sec:requirements} to several other perspectives and directions: cultivating humanoid humanity and subjectivity, building \textit{mind-to-action} mindful and actionable humanoids, supporting transitional omnimodal humanoid modeling, and hybridizing virtual-real humanoid-human-metaverse symbiosis.

\subsection{Addressing Humanoid Humanity Dilemma by Developing Humane Humanoids}
\label{subsec:challenges}
Addressing the humanoid humanity dilemma is  ambitious  with many grand challenges. We briefly discuss the maturity and applicability of existing AI advances to humanoids, the constraints on humanoid systems, and the challenges in developing human-level humanoids. 

First, in addition to issues and challenges, such as hallucinations, of existing generative AI, humanizing humanoids is also challenged by data, modeling, evaluation, ethics, societal and human complexities relevant to humanoids. Typical issues include bias and fairness, distribution shift, contextual understanding, multimodal capabilities, personalized sparse satisfaction, adversarial attacks, misinformation and disinformation, manipulation and deception, observation and data overreliance, overfitting or overmanipulation, lack of thinking and humanity, and potential adoption of technical, security, privacy, ethical, economic, psychological, sociocultural or sociopolitical risks into humanoid systems.

Second, humanoids pose specific constraints and challenges in developing humane and human-level robotic intelligence. There are limitations on humanoid locomotion, mechanics, sensing, responding capabilities, memory, degree of freedom, dimensionality, modality, and computational capabilities, which cannot support LLMs and LMMs on humanoids without edge and cloud networks. More fundamental limitations lie in the existing designs and mechanisms of enabling on-humanoid intelligence such as vision, perception, learning, memorization, control, planning, and actions. Simulating and replicating human intelligence into humanoids requires substantial theoretical and practical advancements in humanoid systems and their cognition, humanity, biomechanics, and behavioral studies, etc. To avoid the uncanny valley phenomena and risks of humanoids and issues like hallucinations of GenAI-driven humanoids, humanoid robotics expects not just human-looking machines but humanness \cite{FongND03} and humanity into humanoids. This requires humane expression, behaving, response, interaction, conservation, and collaboration to make users feel comfortable and safe emotionally and psychologically in interacting and collaborating with humanoids. 

Accordingly, human-level humanoids requires new-generation AI and human-level humanoid AI customized for humanoids toward developing human-like robotic intelligence before aiming for any superintelligence or superhuman intelligence \cite{shin2023superhuman}. Tangible opportunities include developing 1) on-humanoid encoding or decoding-based LLMs, VLP, VLM and LMMs; 2) humanoid-edge-integrative (local server or multiple humanoid networks) LLMs and LMMs; 3) partial humanoid humanity; 4) human-like humanoid intelligent capabilities such as inspired by children learning procedure, and human personality, intentionality, subjectivity and empathy under positive, negative and neutral scenarios and activities; 5) human-like social humanoid intelligent capabilities such as team coordination, cooperation, competition, and conflict resolution during social interactions.

\subsection{Cultivating Humanoid Humanity and Subjectivity}
\label{subsec:humanity-subjectivity}
Without humanity and subjectivity, humanoids are only machines with issues like uncanny valley effect and the humanoid humanity dilemma. Characterizing humane and subjective traits, attributes, qualities and states of humanoids is fundamental for being humankind, human-like to human-level \cite{Cao24}. Humanoid humanity can be embodied through humane attributes such as personality, morality, intentionality, compassion, sentiment and emotion, enabling and managing responsible, trustworthy, sympathetic and ethical individual and collective disposition, behaviors and decisions by humanoids. 

Existing work on embodying such humane and subjective attributes, traits and qualities into humanoids is very limited but focuses on facial expressions, perception and observations. Examples are human mood modeling, such as visual facial expression recognition, textual semantics such as by topic modeling and sentiment analysis, and human body state modeling such as posture recognition \cite{Perez-OsorioW19,OReillyGSW21}. They are observational. Embodying and modeling humanoid humanity and subjectivity requires new thinking and theories beyond the existing humanoid design enabled by observations, such as modeling cognition and mental processes discussed in Section \ref{subsubsec:brain}. 

Fundamentally, a theory of humanoid humanity and subjectivity is essential. This include designing humanoid humanity models, quantifying humanoid subjectivity, structuring humanoid trustfulness, simulating humanoid subjectivity, benchmarking humanoid subjectivity, and developing rules of law for humanoid humanity and subjectivity.
\begin{itemize}
    \item Characterizing humanoid humanity: characterizing and learning humanoid personality, intentionality, consciousness, empathy and trust in real time and live scenarios; defining, imitating and simulating attributes e.g. extroversion, conscientiousness, neuroticism, aggressivity, emotion and intention; and quantifying them in terms of multimodal vision-to-language and perception-to-action tasking, or self-motivated empathy etc. 
    \item Structuring humanoid trust and responsibilities: defining, specifying and simulating the trustfulness and responsibilities of humanoids by humans and humanoid perception and confidence in humans during humanoid and human interaction and collaboration; modeling trust \cite{YeYD23} and responsibility in context of perception-to-action pretraining and finetuning, e.g., quantifying humanoid’s LLM and perception-to-action performance with performance measures and criteria such as robustness, reliability, accuracy, and user satisfaction; formulating humanoid’s self-regulation and self-assessment capability and mechanisms, e.g., accountability, interpretability, and exception resolution; exploring symbiosis between external regulatory agents and humanoids for humanoid ethical assurance, and functional and nonfunctional regulation and governance. 
    \item Humanoid humanity simulation: creating use cases, strategies and simulation testbeds for mimicking, generating, implementing, monitoring and adjusting humanoid personality, intentionality, empathy and trust; simulating humanity dynamics and evolution over interactions and collaborations; simulating and controlling subjective expectations under scenarios such as self-improving emergence, immoral and irrational actions, free of risk and compliance regulation. 
    \item Benchmarking humanoid humanity and subjectivity: developing benchmarkable database, use cases, testbed and checklist for experimentation and case studies of humanoid personality, intentionality, consciousness, empathy and trust; exploring open platforms for uncertain, diversified and large-scale scenarios such as self-improving emergence and hallucination, sampling and algorithmic biases, and irreparable mistakes. 
    \item Rules of law for humanoid humanity management: discovering and summarizing general principles, rules of law for monitoring, detecting, predicting, intervening and managing personality, intentionality, consciousness, empathy and trust of real-time, interactive and multimodal humanoids; identifying verifiable rules and code of conduct for different use cases, such as humanoid tasking, humanoid-human teaming and interactions. 
\end{itemize}

\subsection{Mind-to-Action for Mindful and Actionable Humanoids}
\label{subsec:mind-action}
The concepts of AI mind, mindful AI with AI mindfulness, and AI action, actionable AI and AI actionability \cite{Cao22j} apply to humanoid AI as well, aiming for producing mind-to-action-embodied mindful and actionable humanoids. Humanoid mind modeling is essential for enabling cognitive humanoids and cognition-to-action translation and transformation.

First, making humanoids human-like and humane requires developing humanoid mind and mindful humanoids. A \textit{mindful humanoid} with AI mindfulness simulates and amalgamates human mind and mindfulness to develop \textit{humanoid mind models} and \textit{humanoid mindfulness}. Mindful humanoids may fulfill relevant human thinking mechanisms, cognitive and psychological traits, sensation, reasoning, and critical analysis capabilities. They are also mindful of their personality, intentionality and emotion while pursuing goals, plans, activities and effects. 

Second, humanoids undertake various actions relating to their mind, goals, tasks and environments. \textit{Actionable humanoids} not only conduct such actions but also make them meaningful and valuable technically and practically, i.e., making humanoids actionable. The \textit{actionability} \cite{CaoLZ07} of humanoids, including their designs, actions, and performance, satisfies various performance measures, e.g., interpretable to end users, convertible to business value and impact, and satisfying business, social and economic measures, as outlined in Section \ref{subsec:nonfunctional}. 

Further, a humanoid as a closed-form independent AI system needs the transformation from mind to action. \textit{Humanoid mind-to-action} enables a real-time translation and transfer from subjective mindfulness to actionable operations and activities; the reflection and feedback on actions to mind; and the iterative processes between mind and actions over a task period. This mind-to-action translation raises many research challenges and opportunities, such as building mind models for humanoids, and enabling mind representation (e.g. decomposed to intention representation, sentiment and emotion representation), dynamic and interactive alignment between mind and actions, and coordination between multiple thoughts and their corresponding actions. 

Mindful humanoids are cognitive, replicating and tailoring human cognitive intelligence, traits and properties into humanoid cognitive models. Humanoid cognition can be characterized by representing humanoid's intention, affection, emotion, speech, gesture, actions, haptic signals, and physiological signals. The mind-to-action translation demands \textit{cognition-to-action} modeling and translation, integrating relevant modeling tasks such as personality modeling, intention learning, goal learning, emotion learning, action learning, and behavior learning, and translating learned knowledge on humanoid cognition to their actions.

\subsection{Transitional Omnimodal Humanoid Modeling}
\label{subsec:percept-beh}
Humanoid mind-to-action builds on the quantification of humanoid subjectivity (e.g. personality, intention, sentiment and emotion) and the development of humanoid mind models. It also depends on the fitting, alignment, reasoning,  inference and matching of omnimodal mind-to-action translation \cite{Cao24}. This requires further studies on transforming and translating from vision-language modeling and vision-to-action modeling into mind-to-action modeling and perception-to-behavior modeling.

The \textit{perception-to-behavior} transition, translation and transformation involve multiple to omniversal modalities from visual, acoustic and textual to behavioral aspects. It further raises issues including multimodal to omnimodal perception fusion by synergizing visual, vocal, textual and behavioral representations; point-based, sequential, spatial or function-oriented fitting, alignment and matching between image (or video), sound and text for the same task or object; and temporal, spatial or spatiotemporal developments of multimodal perceptions and behaviors. 

In recent years, an increasing attention has been paid to multimodal deep learning and specifically LMMs, such as visual query-answering (VQA), VLMs, VLP, image-language pretraining, image-to-text generation, and text-to-image generation. Building on Google Robotic Transformer 1 (RT-1) 
on multitask modeling \cite{GoogleRT-1}, Google Robotic Transformer 2 (RT-2) 
\cite{GoogleRT-2} further supports high-capacity VLA modeling on web data and robotic vision data, and translates the learned knowledge into generalized instructions to control robotic actions. RT-2 uses the pathways language and image model (PaLI-X) and pathways language model embodied (PaLM-E) as the backbone.

\subsection{Virtual-Real Humanoid-Human-Metaverse Symbiosis}
\label{subsec:metaverse-humanoid}

\subsubsection{Humanoid Digital Twins and Metaverse} 
While the integration of humanoids with digital twins and metaverse is an open area yet explored, it can nurture novel humanoid animation, humanoid societies, or even  humanoid-human coexisted digital twins or metaverse continuum. By integrating techniques including humanoid robotics, LLMs and LMMs, digital twins, metaverse, decentralized AI, Web3, 3D design, and operating and software systems, humanoid digital twins and metaverse can augment humanoids, promote bidirectional humanoid virtuality-reality transfers, and cultivate virtuality-reality-symbiotic humanoid-human-coexisted metaverse ecosystems. 

First, humanoid digital twins and metaverse surpass humanoid capabilities and capacity. Metaverse, digital twin, and virtual, augmented and mixed reality techniques can assist human-humanoid interaction, interfacing and collaboration in creating metaverse-humanoid-human collaborative design, development, workplace and tasking settings \cite{SuzukiKXHM22,Chang20,BarattaCLN24,Walker23}. On-humanoid, on-body or on-environment animation, metaverse and mixed reality can augment humanoid's collaborative programming, teleoperation in projected environment and background, navigation trajectory visualization, immersive and expressive perception and telepresence. 
For example, by creating a 3D presentation of a humanoid and its workplace, a humanoid can be operated, monitored or tutored for specific tasks in its working environment \cite{Ishiguro2024d}. Instructors with multi-modal virtual, augmented and mixed reality devices can operate and control humanoids and conduct various  tasks, such as imposing humanoid movement restrictions, trajectory teaching, changing humanoid joint angles, and creating task-specific programs. For human-unfriendly environments, humanoids can be teleoperated to undertake specific tasks remotely using projection mapping to facilitate robotic tasks \cite{DarvishPRCPYIP23}.

Second, by integrating with simulation platforms and creative technologies, humanoids can be teleoperated in a digital twin. Initial physical scenes (such as medical test) could be captured and analyzed in the physical world, then converted into the virtual world (e.g., input into 3D simulators) for further manipulation and analysis. Optional actions (e.g., medical treatments) could be tested in the virtual world. Those verified by professionals (e.g., a doctor) could then be deployed to humanoids for actions (e.g., guiding a patient's teletreatment). 
For example, a cybernetic avatar \cite{Ishiguro2024d} is a teleoperated human-like and life-like alter ego of its human operator. Further, in a virtualized real world, a remote operator, its avatar, and a virtual agent (mapping of the avatar) may co-exist in a symbiotic society, where the operator manipulates the virtual agent, which further instructs the avatar to behave and perform like the operator. The avatar mimics facial expressions, eye gaze, gesture, spoken voice, and even mental state and personality of the operator, which are transited through the virtual agent in the avatar online operating platform.

Last but not least, humanoid digital twins and metaverse can transform digital innovations, such as digital health, finance, marketing, customer services and disaster management \cite{CaoDeAI,Ishiguro2024d}. For example, a \textit{humanoid general practitioner} (HGP) may offer online telehealth services as follows \cite{Cao24}. Once receiving the appointment request by a patient, the HGP may launch its virtual digital twin or metaverse platform, where a virtual digital human  illustrates, inquires and discusses health or medical concerns and symptoms with the patient. The patient can tell her problems, upload her medical test results to the metaverse. The HGP can then extract, process and analyze the uploaded materials, animate, illustrate and explain the patient's problems through the digital twin mapped to the patient per her demographics and medical conditions, and make assessment of the patient's complaints. Further diagnosis and treatment suggestions may be made and explained to the patient using the digital human. All discussion, materials and HGP diagnosis and analysis can be sent to a human doctor for verification, adjustment or confirmation. This HGP with health digital twin could save time, extend coverage, and augment service depth and health education  of healthcare and telehealth. A similar example for digital finance would be humanoid robo-advising integrating humanoids with metaverse.

\subsubsection{Humanoid Animation, Imitation and Simulation to Operation}
Simulation, imitation and animation can support humanoid virtuality-to-reality and augment humanoid capacity, operations and performance. 

\textbf{Humanoid Animation and Simulation to Operation}
Large-scale humanoid experiments and practices are costly, risky, insecure, unsafe or unactionable. \textit{Humanoid animation, simulation and programming} provide theories and tools to mimic, evaluate, adjust, optimize and program humanoids in a virtualized cost-effective platform before their confirmation, production and deployment \cite{MuratoreRTYGP22,Ishiguro2024d}. Robotic animators and simulators may provide interfaces, methods and functions to describe, select, connect, code, expand and manage humanoid goals, types, parts, coordinates, objects, environments, functions, tools, workflow, paths and behaviors in a 2D or 3D animation or simulation platform. They can simulate physical systems, activities and processes and generate, compile and deploy animated, simulated or optimized programs into physical robots for further experiment, prototyping, production or deployment. Humanoid animation and simulation may clone, virtualize, imitate or replicate human body parts, attributes, behaviors, emotions, conversations and expressions, and generate humane robotic parts, properties, emotions and actions for more human-like appearance, presentation, and mechanisms. Further, humanoid learning, planning, control and interaction by animation and simulation emerge as essential mechanisms for humanoid robotics at scale.
Humanoid animation, simulation and programming may face \textit{simulation-to-reality gaps} (Sim2Real) from real-life humanoid appearance, design, manipulation and experiment. This gap may relate to misalignment between simulation and reality, and the unrealisticness and poor anthropomorphism of simulation assumptions and constraints on simulation settings,  platforms, essential and sufficient toolsets, programming tools, and simulator behavior settings, capabilities and capacity.

\textbf{Humanoid Imitation to Operation} 
\textit{Imitation learning} is another type of simulation or demonstration. Humanoids learn by demonstrations or are taught by a supervisor, a demonstrator or examples, such as from virtual simulations or a filmed physical scenario (a video), and then clone, imitate or adjust the demonstrations or examples. Humanoids may also learn an underlying optimal reward function or policies from teachers, adapt to  taught examples, develop or generate new behaviors \cite{TagliabueH24}. Humanoid imitation thus supports behavior cloning, intent learning, behavior or scene prediction, recommendation or generation. By integrating techniques like DeepFake, animation, LLMs and digital twins, humanoid imitation could learn from sequential, hierarchical, multimodal and live human activities, and social activities; processes, mechanisms and strategies for handling complex tasks; general and specific skills for problem-solving; and intention, morality and humanity exhibited during human actions and interaction. Various extensions include learning by imitation, learning to imitate \cite{Sun23}, generation by imitation, domain adaptation or domain transfer.  
Humanoid imitation also face issues like imitation-to-reality gaps, in particular, for unseen, new and open world objects, scenarios and tasks, and efficient and effective in-distribution, in-domain and in-context imitation and imitation at scale.

\subsubsection{Humanoid Demonstration to Generation}
Humanoid intelligence can also be learned, augmented and generated by demonstrations and generation, such as by GenAI and advanced models like LLMs, VLM and LMMs. 

\textbf{Humanoid Demonstration-to-Operation}
\textit{Humanoid demonstration} illustrates robots with expected functions, tasks, scenarios, behaviors, actions, and processes. Humanoids can then learn from single, few or sequential demonstrations to operate, plan, navigate or evolve over existing, new, unseen, open or lifelong tasks or environments \cite{ArgallCVB09}. Humanoids may learn from human demonstrations \cite{GoogleRT-2}, such as positioning, posture, behaving, emotional, vocal and facial expressions, and ethical responses, to improve, evolve or develop their world knowledge, human-like actions and responses. Humanoids may also clone, optimize and expand the learned knowledge, activities or intelligence from demonstrations and further forecast and generate new actions, knowledge or intelligence through generative AI.
Humanoid \textit{learning by demonstration} leverages humanoid simulation and mitigates sim-to-real gaps by enhancing humanoid's familiarity and realisticness of the world and building robot experience, memory and reasoning. 
Similar to humanoid simulation, humanoid demonstration faces \textit{demonstration-to-reality gaps} (Demo2Real gaps for short). Such gaps may be caused by demonstration biases, distribution shift, misalignment, or the limitation and insufficiency of demonstrations (e.g., from zero shot, one shot to few shots), settings, capabilities, and capacity. Humanoid learning by demonstration still faces significant challenges for unstructured, ill-structured, unseen-structured, unknown-structured, transitional, lifelong and nonstationary settings and environments. Both humanoid simulation and demonstration are also challenged by zero, new, evolving, unseen, unknown and open objects, contexts, scenarios or tasks.

\textbf{Humanoid Generation-to-Operation}
With the fast-pace development of GenAI including LLMs, LMMs, VQA, VLPs, VLMs and VLAs, \textit{automated generation} becomes feasible and essential in simulating and producing AI and realizing human-level intelligence, substantially improving the intelligence of humanoids. \textit{Humanoid intelligence generation} may present new opportunities and challenges than existing LLMs for image-to-text, video-to-text, text-to-image, and text-to-video, making the generation suitable for humanoid settings. Examples are 1) single-modal generation, e.g., generating humanoid behaviors, actions, responses, and emotions; and 2) cross-modal generation, such as image-to-speech, image-to-conversation, image-to-emotion, animation-to-speech, animation-to-conversation, animation-to-singing, animation-to-dancing, text-to-speech, text-to-conversation, text-to-action, and text-to-emotion \cite{GoogleRT-2,MengYCHMH24}. 
Humanoid generation could be built on humanoid animation, simulation, imitation and demonstration, where animation, simulation, imitation and demonstration serve as supervision, providing one to few shots for robot learning and further generation. Zero-to-few shot learning, weakly-supervised learning, transfer learning, contrastive learning, and Knowledge distillation would be useful to augment humanoid learning and generation by LLMs or LMMs.

\section{Conclusions}
\label{sec:conclusion}

Real-time, interactive, and realistic humanoid robots epitomize the pinnacle of systematic AI development, transcending traditional robotics and intelligent systems. These cutting-edge humanoids represent a new generation of AI robots seamlessly integrating mechanical, electrical, biological, psychological, physiological, ergonomic, aesthetic, and sociocultural technologies and achievements. Leveraging recent AI advancements, particularly in GenAI with LLMs, real-time, interactive, and realistic humanoids demonstrate unprecedented potential in embodying human-like features, senses, behaviors, functions, and intelligences.

State-of-the-art humanoids still fall short of attaining human-like intelligence, although they showcase increasingly impressive AI applications. AI-empowered humanoids are propelling the evolution of human-looking robotics towards human-like and humane elements, features, functions, and qualities. Prior to reaching superintelligence or human-level intelligence, substantial focus will still be on addressing comprehensive functional and nonfunctional requirements, challenges, issues to make humanoids actionable, realistic, and humane.

In addition, while the paper covers a comprehensive overall of the past, present and future of humanoids and humanoid AI, it does not have the capacity to specify and consolidate the relevant techniques in detail and to deepen the technical discussion on functional and nonfunctional specifications. In addition, more technical explorations could be made on how and why to advance humanoids. All of these deserve future substantial efforts over the course of humanoid AI research and development.

\begin{acks}
This work is partially sponsored by the Australian Research Council Discovery grant DP240102050, ARC LIEF grant LE240100131, and ARC Linkage grant LP230201022.
\end{acks}

\bibliographystyle{ACM-Reference-Format}
\bibliography{Humanoid-AI-final}

\end{document}